\documentclass[acmtog]{acmart}
\acmSubmissionID{1431}

\usepackage{multirow} 

\usepackage[ruled]{algorithm2e} 
\SetAlFnt{\small}
\SetAlCapFnt{\small}
\SetAlCapNameFnt{\small}

\begin{document}

\setcopyright{cc}
\setcctype{by}
\acmJournal{TOG}
\acmYear{2026} \acmVolume{45} \acmNumber{6} \acmArticle{244}
\acmMonth{12} \acmDOI{10.1145/3842524}

\title{Direct3D-V3: Latent-Free Diffusion Transformers in the Native Voxel Domain}

\author{Shuang Wu}
\authornote{This work was conducted during an internship at DreamTech.}
\affiliation{%
 \institution{Nanjing University}
 \city{Nanjing}
 \country{China}
}
\affiliation{
  \institution{DreamTech}
  \city{Shenzhen}
  \country{China}
}
\email{wushuang@smail.nju.edu.cn}

\author{Jiachen Qian}
\affiliation{%
 \institution{The University of Hong Kong}
 \city{Hong Kong}
 \country{Hong Kong}}
\affiliation{
  \institution{DreamTech}
  \city{Shenzhen}
  \country{China}
}
\email{qianjiachen@dreamtech.ai}

\author{Youtian Lin}
\affiliation{%
 \institution{Nanjing University}
 \city{Suzhou}
 \country{China}}
\email{linyoutian.loyot@gmail.com}

\author{Feihu Zhang}
\affiliation{%
 \institution{DreamTech}
 \city{Shenzhen}
 \country{China}}
\email{zhangfeihu@dreamtech.ai}

\author{Philip Torr}
\affiliation{%
 \institution{University of Oxford}
 \city{Oxford}
 \country{United Kingdom}}
\email{philip.torr@eng.ox.ac.uk}

\author{Yao Yao}
\authornote{Corresponding authors.}
\affiliation{%
 \institution{Nanjing University}
 \city{Suzhou}
 \country{China}}
\email{yaoyao@nju.edu.cn}

\author{Xun Cao}
\authornotemark[2]
\affiliation{%
 \institution{Nanjing University}
 \city{Nanjing}
 \country{China}}
\email{caoxun@nju.edu.cn}

\begin{abstract}
    Native 3D generation has largely been built around latent diffusion, with recent advances concentrating on stronger variational autoencoders (VAE) and increasingly elaborate latent representations, from vecset and triplanes to structured sparse voxels. Despite their architectural differences, these methods all share the same premise: geometry is first compressed into a latent space and only then modeled by the generator. This paradigm carries two intrinsic limitations. First, VAE reconstruction removes fine geometric detail before the diffusion model ever sees the data. Second, the two-stage pipeline creates an objective mismatch, since the VAE is trained for reconstruction whereas the diffusion model is trained for distribution matching. We take a different route and propose Direct3D-V3, a latent-free diffusion transformer (DiT) framework that discards the VAE altogether and operates in the native voxel domain, directly modeling raw SDF voxel grids during both training and sampling. To make this direct voxel formulation geometrically reliable, we introduce an implicit manifold regularization over neighboring voxels to suppress non-watertight surfaces, and convert predicted SDF grids into meshes via Marching Cubes so that rendered normal losses can supervise the model directly from its predicted SDF. To improve alignment between the conditioning image and generated geometry, Direct3D-V3 also predicts the camera parameters of the conditioning image as an auxiliary task, encouraging the DiT to learn the mapping from image observations to 3D shape. This latent-free design removes the reconstruction bottleneck, brings training under an objective in the native voxel domain, and yields meshes with substantially richer local detail. The predicted camera and native SDF voxel output further make Direct3D-V3 extensible to generative reconstruction, where diffusion is interleaved with normal-guided refinement to improve alignment with the observed view while the generative prior completes unseen regions. Extensive experiments show that Direct3D-V3 achieves competitive results to state-of-the-art latent-based methods, while producing finer geometry. Project page: \href{https://dreamtechai.github.io/Direct3D-V3}{https://dreamtechai.github.io/Direct3D-V3}.
\end{abstract}

%
%


%
%


\begin{teaserfigure}
  \includegraphics[width=\textwidth]{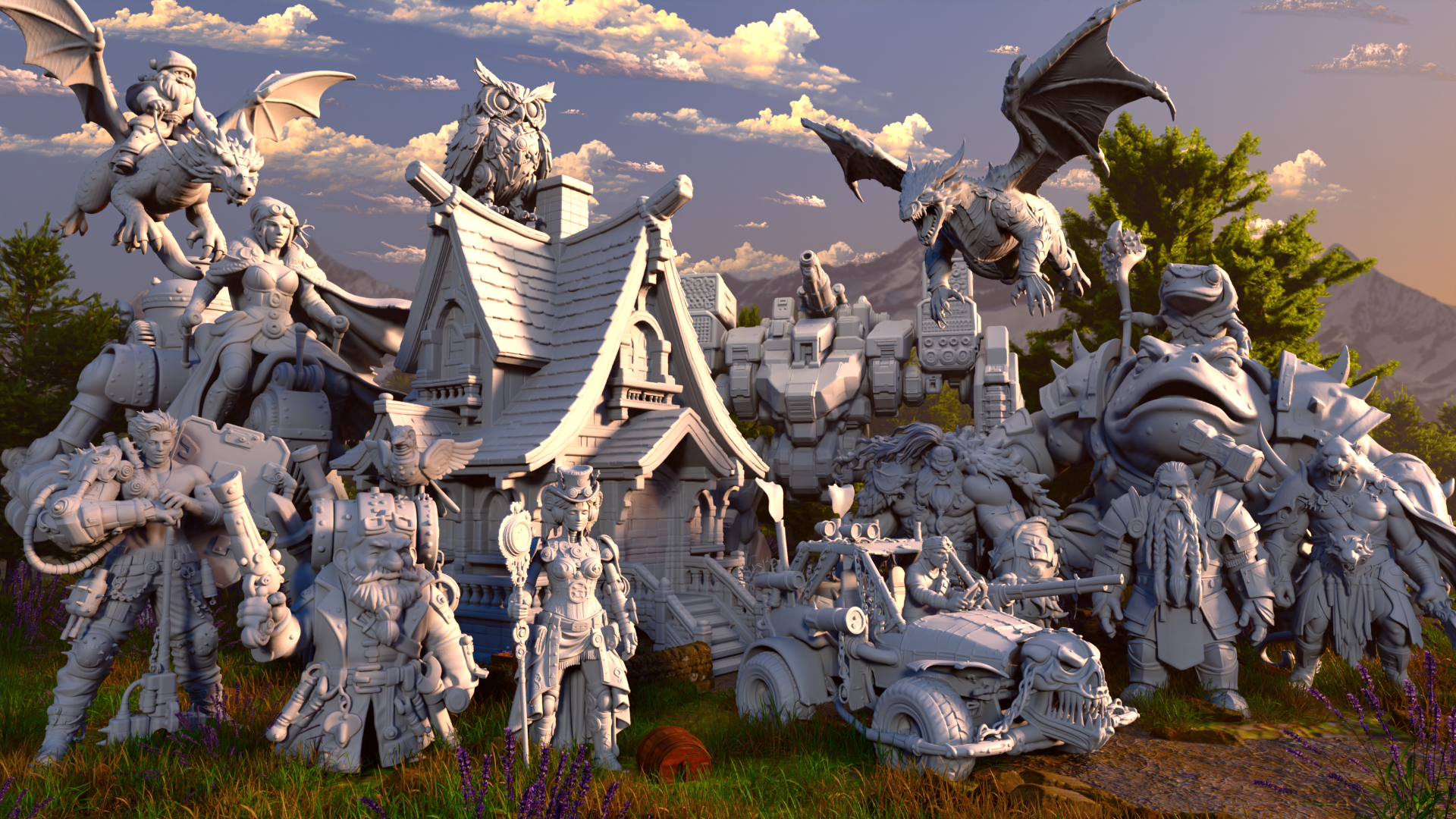}
  \caption{Direct3D-V3 generates diverse, high-fidelity 3D meshes by modeling raw SDF voxel grids directly with latent-free diffusion transformers, removing the VAE bottleneck used by latent diffusion pipelines.}
  \label{fig:teaser}
  \Description{}
\end{teaserfigure}
  
\maketitle

\section{Introduction}
\label{sec:intro}

Recent years have seen rapid progress in 3D shape generation, driven by advances in generative modeling and large-scale 3D datasets. This progress has opened up a wide range of applications, from game and film asset creation to AR/VR content authoring and robotic simulation. In many of these settings, 
the practical goal is to recover a high-fidelity 3D mesh from reference images while preserving faithful geometry, fine surface details, and alignment to the observed view.

The dominant recipe for this problem is latent diffusion~\cite{rombach2022high}. A variational autoencoder (VAE) is first trained to compress 3D shapes into a latent space, and a diffusion model is then trained to generate within that space. 
Following this recipe, recent work has largely focused on how to parameterize the VAE latent space. Existing designs can be broadly organized into vecset representations~\cite{zhang20233dshape2vecset,zhang2024clay}, triplane~\cite{wu2024direct3d}, and structured sparse-voxel latents~\cite{li2025sparc3d,xiang2025structured,wu2025direct3ds2}. Although these designs encode geometry differently, they share the same premise: geometry is first compressed into a learned latent representation and only then modeled by the generator.
This latent paradigm carries two limitations for high-fidelity geometry generation. First, VAE reconstruction acts as an information bottleneck: fine details lost during encoding cannot be recovered by the downstream diffusion model. Second, the two-stage pipeline introduces an objective mismatch, because the VAE is optimized for reconstruction whereas the diffusion model is optimized to match the learned latent distribution. The generator is therefore constrained by both the fidelity of the learned representation and the gap between independently trained objectives.
These observations suggest that further refinement of latent design is unlikely to remove either limitation, and motivate us to revisit the premise of latent compression itself.

We therefore take a different route and propose Direct3D-V3, a latent-free diffusion transformer (DiT) framework that operates in the native voxel domain by directly modeling raw SDF voxel grids during both training and sampling.
This places the generator in the same signal space where geometry is ultimately evaluated, but also makes scalability the central challenge: a dense SDF volume grows cubically with resolution. To address this, we adopt a dual-granularity architecture: a patch-level DiT reasons about coarse semantics and global layout on a sparse sequence of patch tokens, while a lightweight voxel-level DiT restores fine geometric detail at voxel granularity. This coarse-to-fine organization reflects a broader principle in high-resolution pixel-space diffusion~\cite{wang2025ddt,yu2025pixeldit}: global structure and high-frequency detail are easier to model when they are handled at compatible granularities. For 3D geometry, surface sparsity makes this principle particularly effective, since the most informative signed-distance variation lies near the zero-level set.
Concretely, the patch-level model operates on surface patches, while the voxel-level model predicts high-frequency details within selected patches.
Together, these modules make native SDF diffusion tractable at \mbox{$1536^3$}.

Operating directly on SDF voxels introduces a new optimization challenge, but also opens up a new form of geometric supervision. On the one hand, even small perturbations in predicted SDF values can cause Marching Cubes to extract non-watertight surfaces with open boundaries, a failure mode that VAE decoders tend to suppress implicitly by learning the manifold structure of meshes. We address this issue with an implicit manifold regularization loss that penalizes abrupt SDF discontinuities between neighboring voxels, substantially reducing non-watertight artifacts in practice. On the other hand, once the DiT predicts raw SDFs, each grid can be converted directly into a mesh via Marching Cubes, allowing rendered normal losses to supervise the SDF voxel output through the extracted mesh. A related idea was explored by Hi3DGen~\cite{ye2025hi3dgen}, but its reliance on VAE decoding for normal rendering incurs substantial memory overhead, limiting its applicable resolution to $256^3$. By contrast, our approach operates without such a bottleneck, scaling to resolutions as high as $1536^3$, and the resulting normal supervision yields measurable improvements in both the visual quality and geometric fidelity of the generated meshes. To improve alignment between the conditioning image and generated geometry, Direct3D-V3 also predicts the camera parameters of the conditioning image as an auxiliary task, encouraging the DiT to learn the mapping from image observations to 3D shape.

With the predicted camera and native SDF voxel output, Direct3D-V3 can be extended to generative reconstruction. The camera supplies the input view for test-time refinement, while the raw SDF grid provides geometry that can be optimized without repeatedly passing through a heavy VAE decoder. Building on this interface, we first use a pre-trained normal prediction model to estimate the normal map of the input image. We then extract a mesh from the predicted SDF via Marching Cubes, render the mesh normal map under the predicted camera, and optimize the discrepancy between the rendered normal and the predicted normal. We interleave this normal-based refinement with diffusion sampling, allowing the reconstruction signal to tighten alignment with the input image while the generative prior continues to complete unseen regions and regularize the surface.

In summary, the main contributions of this paper are:
\begin{itemize}
    \item We propose Direct3D-V3, a latent-free DiT framework that trains and samples directly on raw SDF voxel grids in the native voxel domain and scales to a resolution of $1536^3$ through a dual-granularity architecture.
    \item We equip the native voxel formulation with three complementary signals: implicit manifold regularization for SDF consistency, rendered normal supervision for surface detail, and an auxiliary camera branch for alignment between the conditioning image and generated geometry.
    \item We further show that the latent-free design extends naturally to generative reconstruction, enabling effective normal-guided test-time refinement while preserving the geometric prior learned by the generator.
    \item Extensive experiments on image-to-3D generation show that Direct3D-V3 achieves competitive or superior performance to state-of-the-art latent-based methods, while producing meshes with finer geometric detail and better alignment with the input image.
\end{itemize}

\section{Related Work}
\label{sec:related-work}

\begin{figure*}[!t]
  \centering
  \includegraphics[width=\linewidth]{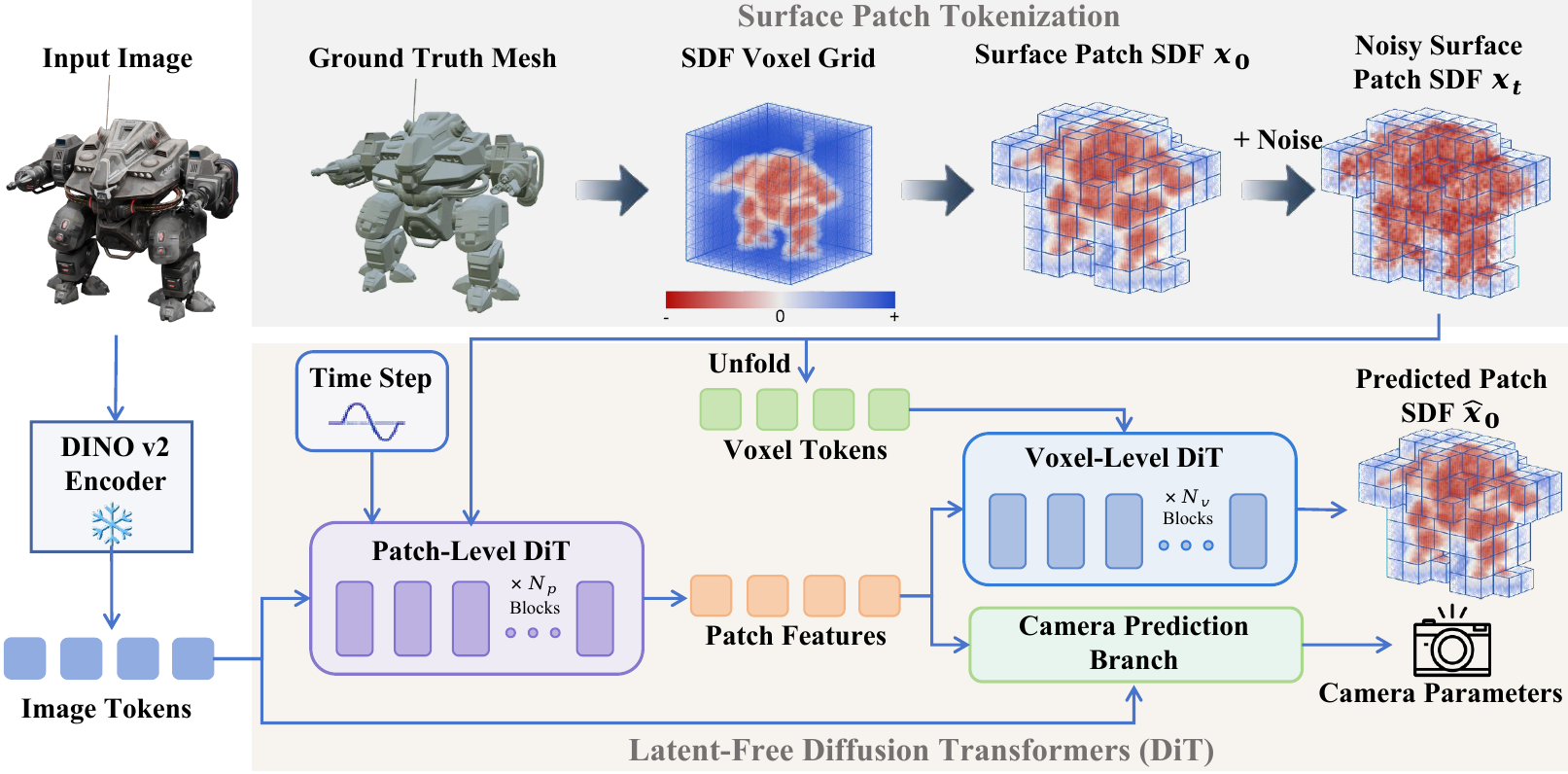}
  \caption{Framework of Direct3D-V3. Rather than compressing geometry into a learned latent representation, Direct3D-V3 trains and samples directly on raw SDF voxel grids. During training, we extract the SDF voxel grid and surface patch SDF values from the ground-truth mesh, and add Gaussian noise to the surface patch SDF values to form the DiT input $\mathbf{x}_t$. During inference, the first-stage model is used only to provide the surface patch indices; sampling then starts from Gaussian noise on these sparse patches, and the latent-free DiT predicts the clean SDF values directly in the native voxel domain. The patch-level DiT uses the noisy SDF signal, DINOv2 image tokens, and timestep embedding to reason efficiently over global shape semantics, while a lightweight voxel-level DiT refines voxel-level SDF values within each selected patch. In parallel, a camera prediction branch combines patch-level features with image tokens to regress normalized intrinsic and extrinsic parameters, encouraging the generated shape to remain spatially aligned with the conditioning image.}
  \Description{}
  \label{fig:framework}
\end{figure*}

\subsection{Native 3D Generation}

Recent progress in 3D generation has been dominated by latent diffusion pipelines, in which a variational autoencoder first compresses geometry into a compact latent space and a diffusion model is then trained to generate in that space~\cite{rombach2022high}. Within this paradigm, one important line of work builds on vecset~\cite{zhang20233dshape2vecset} representations, which provide an implicit latent description of 3D shape through a fixed-length set of tokens that can be decoded into neural fields. CLAY~\cite{zhang2024clay} scales this representation to controllable large-scale 3D asset generation. TripoSG~\cite{li2025triposg} further combines vecset latents with large-scale rectified flow training to improve synthesis fidelity. Dora~\cite{chen2024dora} and Hunyuan3D 2.0~\cite{zhao2025hunyuan3d2} study this line from the VAE perspective, showing that sharp edge sampling and stronger cross-attention improve the preservation of fine geometric structures during shape autoencoding.

Another representative family uses explicit triplane features. Direct3D~\cite{wu2024direct3d} factorizes the latent into three orthogonal feature planes, injecting explicit positional structure into the latent representation and making diffusion modeling easier to scale than purely implicit token sets. Hyper3D~\cite{guo2025hyper3d} further augments the triplane latent with a low-resolution 3D grid latent, which supplements explicit 3D information and supports higher resolution representation with more efficient training.

A third and rapidly growing family uses sparse voxel structures as the latent representation. TRELLIS~\cite{xiang2025structured} proposes the unified structured latent representation, which can be decoded into multiple downstream 3D formats, including NeRF~\cite{mildenhall2020nerf}, FlexiCubes~\cite{shen2023flexible}, and 3D Gaussian Splatting~\cite{kerbl20233d}. Direct3D-S2~\cite{wu2025direct3ds2} introduces spatial sparse attention, which significantly improves both training and inference efficiency for high resolution generation. SparseFlex~\cite{he2025sparseflex} focuses on high resolution shape modeling with arbitrary topology, extending sparse structured generation to more flexible geometric outputs. Sparc3D~\cite{li2025sparc3d} introduces a sparse deformable marching cubes representation that can efficiently convert raw meshes to high resolution watertight ones. TRELLIS2~\cite{xiang2026native} introduces a new sparse voxel structure, O-Voxel, that can jointly encode geometry and appearance in a compact native representation. Despite their differences in architecture and engineering, all of these methods still rely on compressing geometry into a learned latent space before generation. Our work departs from this common assumption and asks whether the latent stage can be removed entirely.

\subsection{Latent-Free Diffusion Models}

The idea of training diffusion models directly in data space has deep roots in image generation. Early models such as DDPM~\cite{ho2020ddpm}, Improved DDPM~\cite{nichol2021improved}, ADM~\cite{dhariwal2021adm}, and EDM~\cite{karras2022edm} all operate directly in pixel space and establish the effectiveness of end-to-end denoising in the original signal domain. As model scale increased, however, many works shifted to latent space for efficiency. LDM~\cite{rombach2022high} made latent compression the dominant recipe for high-resolution diffusion, and transformer-based generators such as DiT~\cite{peebles2023dit} and U-ViT~\cite{bao2023uvit} further strengthened this trend by showing that large transformer backbones remain highly effective when operating on compact latent tokens.

Recent work has returned to the question of whether latent compression is necessary at all. JiT~\cite{li2025back} argues that diffusion models should directly predict clean data rather than noise-like quantities, and shows that simple large patch transformers can already become strong pixel-space generators without tokenizers or auxiliary losses. DiP~\cite{chen2025dip} studies how to make pixel-space diffusion practical at higher resolution by combining a DiT backbone with a lightweight patch detailer head, balancing global structure modeling with local detail synthesis. DeCo~\cite{ma2025deco} observes that modeling low-frequency semantics and high-frequency details with a single transformer is inefficient, and therefore decouples them using a semantic DiT, a lightweight pixel decoder, and a frequency-aware flow matching objective. PixelDiT~\cite{yu2025pixeldit} instead adopts a fully transformer-based dual-level design, where a patch-level DiT captures global semantics and a pixel-level DiT restores texture details directly in pixel space. These methods show that latent-free diffusion can be competitive for image generation, but they cannot be directly transferred to 3D generation, where SDF voxel grids are sparse, memory-intensive, and coupled to mesh extraction. SDF-Diffusion~\cite{shim2023sdfdiffusion} provides an early 3D counterpart by avoiding VAE compression, yet its generation is limited to $128^3$ resolution, which restricts the geometric precision of the resulting shapes. Our Direct3D-V3 extends the latent-free principle to high resolution native 3D generation by operating on surface SDF voxel grids and scaling to resolutions such as $1536^3$.

\subsection{Generative Reconstruction}

One important line of work uses image diffusion priors to improve image-to-3D reconstruction. Zero-1-to-3~\cite{liu2023zero123} learns viewpoint-conditioned novel view synthesis from a single image and shows that a diffusion prior with explicit relative camera control can already support zero-shot 3D reconstruction. SyncDreamer~\cite{liu2023syncdreamer} improves this direction with a synchronized multi-view diffusion model that generates all target views jointly and correlates them through 3D-aware feature attention, substantially improving cross-view consistency. Wonder3D~\cite{long2023wonder3d} predicts both multi-view color images and multi-view normal maps through cross-domain diffusion, then fuses these geometric cues into textured mesh reconstruction. InstantMesh~\cite{xu2024instantmesh} combines an off-the-shelf multi-view diffusion model with a sparse-view large reconstruction model and a differentiable iso-surface extraction module, enabling high-quality mesh generation in only a few seconds. These methods significantly improve coarse structure recovery, novel view consistency, and surface appearance, but their optimization is still driven primarily by image-space consistency rather than direct alignment of fine geometry with the observed input view.

Another related line exploits video diffusion priors for view synthesis and reconstruction. Stable Virtual Camera~\cite{zhou2025stable} formulates generative view synthesis as a general diffusion model conditioned on arbitrary input views and target camera trajectories, producing temporally smooth novel views under large viewpoint changes. GEN3C~\cite{ren2025gen3c} further introduces a 3D cache built from depth predicted on seed images or previous frames, and conditions video generation on renderings of this cache under user-specified camera trajectories. This design yields more precise camera control and stronger temporal 3D consistency than prior video diffusion models. These methods offer richer camera-aware priors than single-image diffusion, but their supervision still lives mainly in image or video space and is not directly tied to mesh-level geometric detail.

Closest to our setting are methods that place reconstruction constraints inside generation itself. ReconViaGen~\cite{chang2025reconviagen} designs global geometry conditions for coarse structure, local per-view conditions for fine geometry and texture, and rendering-aware velocity compensation to correct the denoising trajectory toward the input views. CUPID~\cite{huang2025cupid} uses a two-stage flow pipeline: it first predicts an occupancy cube and a UV cube to estimate the object-centric camera, then uses a pose-aligned conditioner to inject pixel-aligned image features for detailed geometry and appearance. Pixal3D~\cite{li2026pixal3d} generates 3D assets in a view-aligned space to better preserve image-to-3D correspondence and input details. Our method follows this direction, but benefits from a simpler native interface: the latent-free DiT directly outputs SDF voxel grids rather than an intermediate latent representation. This allows the generated geometry to be converted into a mesh by Marching Cubes, rendered under the predicted camera, and directly aligned to normal maps estimated from the input view by a pre-trained normal predictor. As a result, our framework can push fine geometric detail to match the input image more directly, rather than focusing mainly on overall structure or texture.

\section{Latent-Free Diffusion Transformers for 3D Generation}
\label{sec:method}

Existing latent-based methods compress 3D shapes into an intermediate latent space, introducing an unavoidable fidelity gap between the original geometry and what the diffusion model ultimately learns to generate. To circumvent this fundamental limitation, Direct3D-V3 performs training and sampling directly on raw SDF voxel grids, eliminating the VAE entirely, as detailed in Section~\ref{sec:latent-free}. To balance geometric expressiveness with computational tractability at high resolutions, Direct3D-V3 employs two complementary DiT modules operating at patch and voxel granularity respectively, decoupling coarse shape semantics from fine-grained surface detail, as described in Section~\ref{sec:dual}. To ensure the geometric validity of generated shapes, we introduce an implicit manifold regularization that enforces smooth SDF transitions between neighboring voxels, detailed in Section~\ref{sec:manifold}. Since the model predicts SDF voxel grids directly, we can also extract meshes with Marching Cubes and apply rendered normal supervision in the geometric space, as detailed in Section~\ref{sec:geo-opt}. Finally, the framework additionally predicts the camera parameters of the conditioning image, which pushes the generated mesh to align more closely with the input view, as elaborated in Section~\ref{sec:camera-prediction}. An overview of the framework is shown in Figure~\ref{fig:framework}.

\begin{figure}[!t]
  \centering
  \includegraphics[width=\linewidth]{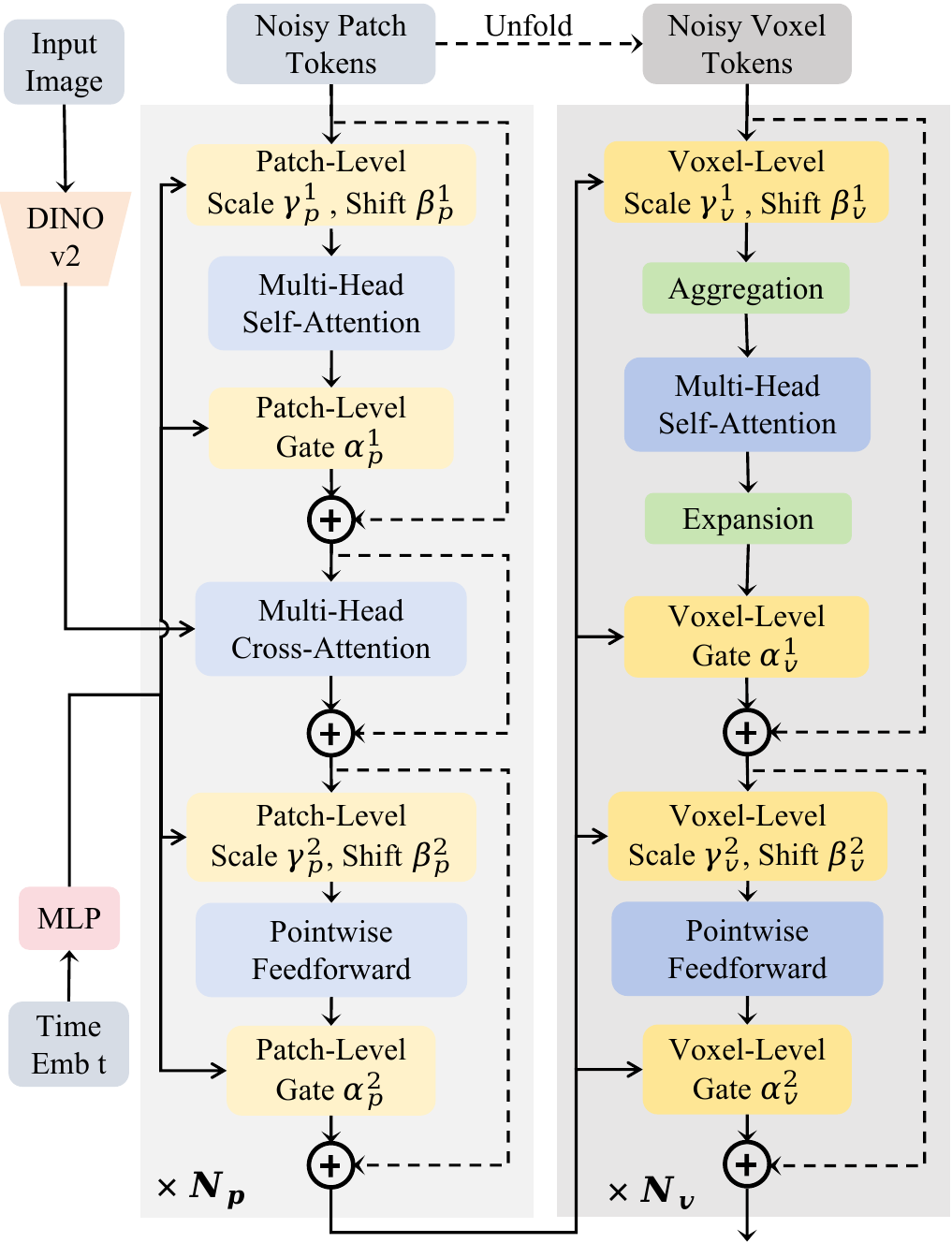}
  \caption{The architecture of the dual-granularity DiT.}
  \Description{}
  \label{fig:dit}
\end{figure}

\subsection{Latent-Free Diffusion on SDF Voxel Grids}
\label{sec:latent-free}

Latent diffusion has become a practical way to reduce the cost of high-dimensional generation, but in 3D shape generation the compression stage also constrains the geometry that the diffusion model can learn. A 3D VAE must encode detailed surface structures into a compact representation and later decode them back to geometry, which can discard fine local detail and introduce the reconstruction bottleneck discussed above. The latent space is also optimized for reconstruction fidelity rather than for the distributional objective used by diffusion, leading to a mismatch between the two training stages. We therefore discard the latent space and train the diffusion model directly in the voxel domain, using raw SDF voxel grids as the generative state.

\textbf{Geometric Formulation on SDF Voxel Grids.} We represent a 3D shape as a uniform SDF voxel grid $\mathbf{X} \in \mathbb{R}^{R \times R \times R \times 1}$, where each voxel stores a SDF value, and R denotes the grid resolution. While our latent-free formulation is in principle compatible with other volumetric representations such as FlexiCubes~\cite{shen2023flexible}, Sparc3D~\cite{li2025sparc3d}, and O-Voxel~\cite{xiang2026native}, we deliberately adopt this minimal representation to validate our core hypothesis that latent-free diffusion can match or surpass latent-based methods without relying on any sophisticated representation design.

\textbf{Surface Patch Tokenization.} Directly operating on a dense voxel grid of resolution $R^3$ is computationally prohibitive for the attention mechanisms in DiT. To reduce the token sequence length, we partition the voxel grid into non-overlapping patches of size $p^3$, yielding $(\frac{R}{p})^3$ patch tokens, each of dimensionality $p^3$. However, even with patch tokenization, the number of tokens at high resolutions such as $1536^3$ remains large. We further exploit the observation that SDF values are only meaningful in the vicinity of the mesh surface, and adopt a sparse surface voxel representation that restricts computation to surface-adjacent patches. Concretely, we follow the two-stage generation paradigm that has become standard practice in high-resolution 3D generation methods such as TRELLIS~\cite{xiang2025structured}, Direct3D-S2~\cite{wu2025direct3ds2}, and LATTICE~\cite{lai2025lattice}. In the first stage, we employ a model to rapidly generate a coarse mesh at low resolution, from which we identify the set of occupied surface patches $\mathcal{S}$.
In the second stage, our latent-free DiT performs diffusion exclusively over these sparse patches, operating on the token sequence 
$\left\{ \mathbf{x}_i \in \mathbb{R}^{p^3} \mid i \in \mathcal{S} \right\}$, 
where $\lvert \mathcal{S} \rvert \ll \left(R/p\right)^3$.
The first-stage model is only used to localize surface patches and is not a focus of this paper. In practice, this stage can be implemented in the same spirit as existing high-resolution multi-stage pipelines: one may generate a coarse mesh with a vecset-based model and extract the corresponding patch indices, or predict the occupied patch set directly with a dense voxel DiT as in TRELLIS. Our method is agnostic to this choice, as long as it provides the surface patch index set $\mathcal{S}$ for the high-resolution latent-free DiT.

\textbf{Native Flow Matching in the Geometric Domain.}
Unlike latent diffusion models that perform denoising in an abstract latent space, our DiT operates natively in the geometric domain, learning to denoise raw SDF voxel grids directly. We train our DiT using the flow matching~\cite{albergo2023building,lipmanflow2023flow} objective, which constructs a probability path between Gaussian noise and the data distribution via the linear interpolation schedule:
\begin{equation}
    \mathbf{x}_t = t \cdot \mathbf{x}_0 + (1 - t) \cdot \boldsymbol{\epsilon}, 
    \quad \boldsymbol{\epsilon} \sim \mathcal{N}(\mathbf{0}, \mathbf{I}),
    \quad t \in [0, 1],
\end{equation}
with ground-truth velocity $\mathbf{v}_t = \mathbf{x}_0 - \boldsymbol{\epsilon}$. Following the $x_0$-prediction paradigm of JiT~\cite{li2025back}, our network $f_{\boldsymbol{\theta}}$ directly predicts the clean voxel grid $\hat{\mathbf{x}}_0$ from the noisy observation $\mathbf{x}_t$ at timestep $t$:
\begin{equation}
    \hat{\mathbf{x}}_0 = f_{\boldsymbol{\theta}}\!\left(\mathbf{x}_t,\, t,\, 
    \mathbf{c}\right),
\end{equation}
where $\mathbf{c}$ denotes the conditioning signal. The predicted 
$\hat{\mathbf{x}}_0$ is converted to the estimated velocity via:
\begin{equation}
    \hat{\mathbf{v}}_t = \frac{\hat{\mathbf{x}}_0 - \mathbf{x}_t}{1 - t},
\end{equation}
and the flow matching loss is defined as:
\begin{equation}
    \mathcal{L}_{\mathrm{fm}} = 
    \mathbb{E}_{t,\, \mathbf{x}_0,\, \boldsymbol{\epsilon}}
    \left[\,\left\| \hat{\mathbf{v}}_t - \mathbf{v}_t \right\|^2\,\right].
\end{equation}

\subsection{Dual-Granularity DiT Architecture}
\label{sec:dual}

While sparse patch tokenization substantially reduces the sequence length for attention computation, it introduces a tension between efficiency and geometric fidelity. A large patch size keeps the number of tokens manageable, but results in patch embeddings of high dimensionality $p^3$, making it difficult for a single DiT to simultaneously model global shape semantics and recover fine-grained 
surface detail. Conversely, operating directly at voxel granularity preserves geometric detail but renders attention computation intractable. Inspired by the frequency decoupling strategy of DDT~\cite{wang2025ddt} and PixelDiT~\cite{yu2025pixeldit}, which separate low-frequency semantic content from high-frequency detail, we propose a Dual-Granularity DiT architecture that decomposes this challenge across two complementary modules: a patch-level DiT that models coarse shape semantics and global layout at patch granularity, and a lightweight Voxel-level DiT that restores fine-grained geometric detail at voxel granularity, conditioned on the output of the patch-level DiT. The architecture is illustrated in Figure~\ref{fig:dit}.

\textbf{Patch-Level DiT.}
The Patch-level DiT, which comprises $N_p$ DiT blocks, operates on the patch token sequence 
$\left\{ \mathbf{x}_i \in \mathbb{R}^{p^3} \mid i \in \mathcal{S} \right\}$ and is responsible for modeling the semantic structure and global layout of the generated shape. Within each block, AdaLN injects the diffusion timestep embedding \mbox{$t$} by predicting global modulation parameters \mbox{$(\gamma_p^1,\beta_p^1,\alpha_p^1,\gamma_p^2,\beta_p^2,\alpha_p^2)$}, which jointly normalize and modulate the patch-token features during denoising. A cross-attention layer incorporates the image conditioning $\mathbf{c}$, extracted by a DINOv2-Large~\cite{oquab2023dinov2} encoder, and a self-attention layer models the dependencies among patch tokens, capturing the global structural relationships of the 3D shape. The output of the patch-level DiT is a sequence of patch features $\mathbf{s}_p \in \mathbb{R}^{L \times D_p}$, where $L = \lvert \mathcal{S} \rvert$ denotes the number of sparse patches and $D_p$ is the hidden size.

\textbf{Voxel-Level DiT.}
While the patch-level DiT captures global structure efficiently, its large patch granularity makes it difficult to preserve fine geometric variation inside each patch. We therefore introduce a lightweight Voxel-Level DiT that refines local surface detail under the guidance of the patch-level output. For the $L$ sparse patches selected by the first stage, the input SDF values to the voxel branch are arranged as $\mathbf{X}_v \in \mathbb{R}^{L \times p^3 \times 1}$. We reshape them into a voxel token sequence of length $L \cdot p^3$ and map each scalar SDF value to a voxel-level embedding $\mathbf{H}^{(0)} \in \mathbb{R}^{(L \cdot p^3) \times D_v}$ through a linear projection, where $D_v$ denotes the hidden dimension of the voxel branch. The patch-level output $\mathbf{s}_p \in \mathbb{R}^{L \times D_p}$ serves as the conditioning signal for local refinement. At the $l$-th Voxel-Level DiT block, a linear layer takes $\mathbf{s}_p$ as input and predicts voxel-level AdaLN parameters
\begin{equation}
    \mathbf{m}^{(l)} = g^{(l)}(\mathbf{s}_p)
    \in \mathbb{R}^{L \times (p^3 \cdot 6 D_v)}.
\end{equation}
After reshaping, these parameters provide the modulation terms for each voxel token, written analogously as \mbox{$[\gamma_v^1, \beta_v^1, \alpha_v^1, \gamma_v^2, \beta_v^2, \alpha_v^2]$} and used together in AdaLN to normalize and modulate each voxel-token feature. In this way, the patch-level feature does not merely provide a single coarse conditioning vector, but modulates all voxel tokens inside the patch at every block. Direct self-attention over all $L \cdot p^3$ voxel tokens is still unnecessarily expensive. We therefore reshape the voxel features back to patch form, aggregate the $p^3$ voxel tokens within each patch into one compact patch token before self-attention, and expand them back afterward:
\begin{equation}
    \begin{aligned}
    \mathbf{U}^{(l)}
    &= \mathcal{A}_{\mathrm{agg}}^{(l)}
    \!\left(\mathbf{H}^{(l)}_{\mathrm{p}}\right), \\
    \overline{\mathbf{U}}^{(l)}
    &= \mathrm{SelfAttn}^{(l)}\!\left(\mathbf{U}^{(l)}\right), \\
    \widehat{\mathbf{H}}^{(l)}_{\mathrm{p}}
    &= \mathcal{A}_{\mathrm{exp}}^{(l)}
    \!\left(\overline{\mathbf{U}}^{(l)}\right).
    \end{aligned}
\end{equation}
where $\mathbf{H}^{(l)}_{\mathrm{p}} \in \mathbb{R}^{L \times p^3 D_v}$ is obtained by regrouping voxel tokens patch by patch. The aggregation module $\mathcal{A}_{\mathrm{agg}}^{(l)}$ maps $\mathbb{R}^{p^3 D_v}$ to $\mathbb{R}^{D_p}$, and the expansion module $\mathcal{A}_{\mathrm{exp}}^{(l)}$ maps $\mathbb{R}^{D_p}$ back to $\mathbb{R}^{p^3 D_v}$; both are lightweight projection modules implemented with linear layers, bias terms, and activation functions. The compact tokens $\mathbf{U}^{(l)}, \overline{\mathbf{U}}^{(l)} \in \mathbb{R}^{L \times D_p}$ attend across sparse patches, and the expanded features $\widehat{\mathbf{H}}^{(l)}_{\mathrm{patch}} \in \mathbb{R}^{L \times p^3 D_v}$ are finally reshaped to $\mathbb{R}^{(L \cdot p^3) \times D_v}$. This aggregation and expansion design keeps the voxel branch lightweight while still allowing the model to restore fine surface geometry under the guidance of patch-level semantics.

Overall, the patch-level DiT provides global semantic structure over sparse surface patches, while the voxel-level DiT restores local SDF variation within each patch. This dual-granularity design avoids applying expensive attention to all voxel tokens directly, yet still preserves the fine geometric detail needed for high-resolution shape generation.

\begin{figure}[!t]
  \centering
  \includegraphics[width=0.95\linewidth]{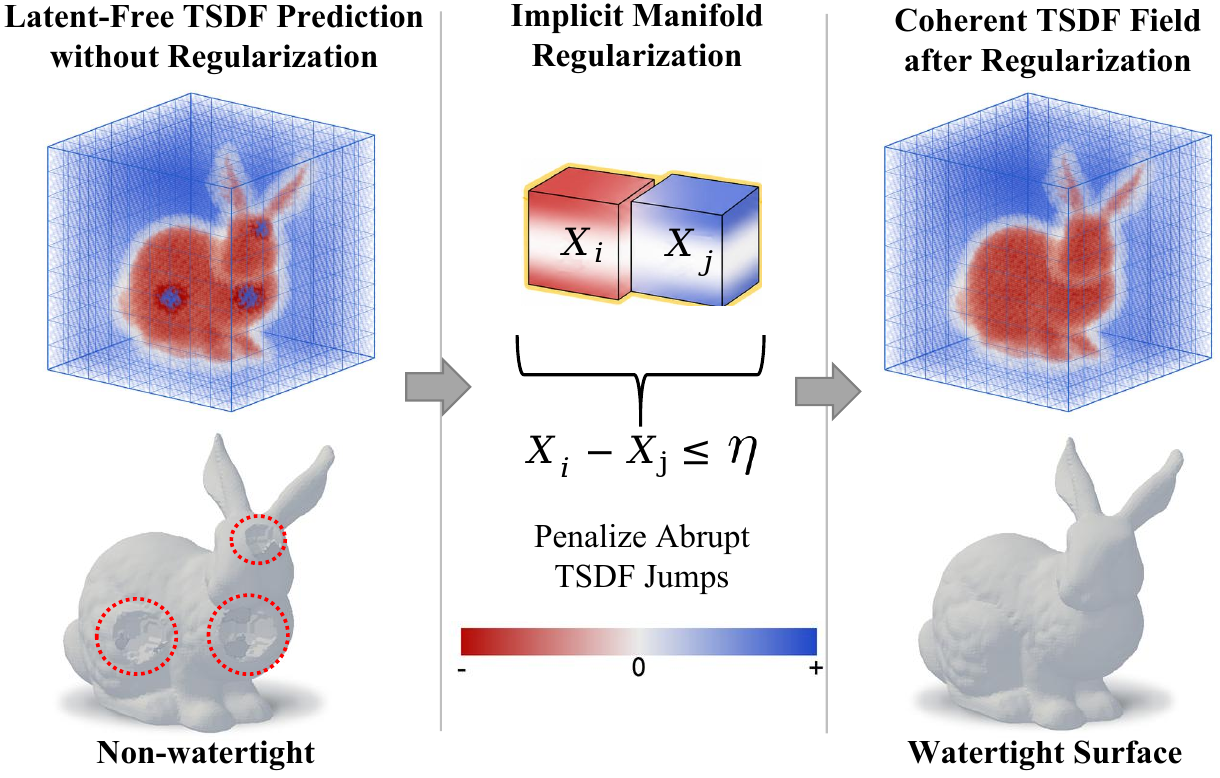}
  \caption{Illustration of the Implicit Manifold Regularization (IMR). Without the decoder-induced geometric prior used in latent-based pipelines, a latent-free DiT can predict locally inconsistent TSDF fields, leading to abrupt zero-level crossings and non-watertight meshes after Marching Cubes. Since the surface manifold is implicitly defined by the zero level set of the TSDF, we regularize this manifold by penalizing neighboring voxels whose predicted SDF values differ by more than one voxel length $\eta$. This local constraint encourages coherent TSDF transitions around the surface, reducing holes and producing a more watertight extracted mesh while preserving the generated shape.}
  \Description{}
  \label{fig:imr}
\end{figure}

\subsection{Implicit Manifold Regularization}
\label{sec:manifold}

Removing the VAE changes not only the representation space, but also the geometric prior available to the generator. In latent-based methods, the VAE decoder is trained to reconstruct valid 3D shapes and therefore implicitly learns the local structure of plausible SDF fields around mesh surfaces. Since the mesh surface is extracted as the zero level set of the SDF, this local distance-field structure acts as an implicit prior on the extracted surface manifold. As a result, even noisy or inaccurate compressed features can often be decoded into a watertight surface without large holes. A latent-free DiT lacks this decoder-induced prior: small local errors in predicted SDF values can create inconsistent zero crossings after Marching Cubes, producing meshes with holes or non-watertight regions. We address this issue with implicit manifold regularization, which does not directly constrain mesh topology, but regularizes the local TSDF field that implicitly defines the surface manifold.

Eikonal regularization is commonly used in 3D reconstruction to constrain learned implicit fields. In our setting, however, the DiT is directly supervised by ground-truth TSDF values, while the Eikonal loss enforces a unit 2-norm gradient at arbitrary spatial points. This constraint can conflict with the empirical TSDF distribution, making DiT optimization unstable and in severe cases causing training collapse. We therefore impose a discrete Lipschitz constraint on neighboring voxels, as illustrated in Figure~\ref{fig:imr}.
Let $\hat{\mathbf{X}}$ denote the predicted TSDF grid and let $\eta$ be the physical length of one voxel. For each pair of neighboring voxels $(i,j) \in \mathcal{N}$, a valid TSDF field should not change by more than one voxel length. We therefore define the implicit manifold regularization loss as:
\begin{equation}
    \mathcal{L}_{\mathrm{imr}}
    =
    \frac{1}{|\mathcal{N}|}
    \sum_{(i,j)\in\mathcal{N}}
    \left[
    \left|\hat{\mathbf{X}}_i - \hat{\mathbf{X}}_j\right| - \eta
    \right]_+^2,
\end{equation}
where $[\cdot]_+ = \max(\cdot,0)$ and $\mathcal{N}$ contains adjacent voxel pairs along the three grid axes. This loss penalizes only abrupt TSDF jumps that violate the local distance-field constraint, while leaving valid local variation unconstrained.

By injecting this implicit manifold prior directly into the latent-free DiT, the model learns to produce smoother and more coherent TSDF fields, which in turn improves the watertightness of meshes extracted by Marching Cubes. We include this term in the final training objective introduced at the end of the method section.

\subsection{Native Geometric Optimization}
\label{sec:geo-opt}

The latent-free formulation enables supervision to be applied directly in the native geometric space. Since the DiT predicts the clean SDF voxel grid $\hat{\mathbf{x}}_0$ rather than an intermediate latent representation, each prediction can be immediately converted into a mesh through Marching Cubes. Given randomly sampled camera parameters $\pi$, we render a normal map from the extracted mesh and compare it with the normal map rendered from the ground-truth mesh under the same camera. This creates a direct geometric supervision signal on the generated surface, rather than supervising the model through a compressed proxy.

Formally, let
\begin{equation}
    \hat{\mathbf{M}} = \mathrm{MC}\!\left(\hat{\mathbf{x}}_0\right),
    \qquad
    \hat{\mathbf{N}}_{\pi}
    = \mathcal{R}_{\mathrm{n}}\!\left(\hat{\mathbf{M}}, \pi\right),
\end{equation}
where $\mathrm{MC}(\cdot)$ denotes Marching Cubes and $\mathcal{R}_{\mathrm{n}}(\cdot,\pi)$ denotes normal rendering under camera $\pi$. In practice, we use differentiable sparse Marching Cubes to extract meshes from the predicted sparse SDF grid, and render normal maps with Nvdiffrast under the specified camera parameters. This pipeline evaluates the supervision on the extracted surface while keeping gradients connected to the native SDF prediction. Let $\mathbf{N}_{\pi}^{\mathrm{gt}}$ be the normal map rendered from the ground-truth mesh with the same camera. We define the native geometric optimization loss as
\begin{equation}
    \mathcal{L}_{\mathrm{geo}}
    =
    \left\|
    \hat{\mathbf{N}}_{\pi}
    -
    \mathbf{N}_{\pi}^{\mathrm{gt}}
    \right\|_1 .
\end{equation}
This loss encourages the predicted SDF grid to produce meshes whose visible surface normals match the target geometry, improving the visual fidelity of generated meshes.

Normal-based supervision has also been explored by Hi3DGen~\cite{ye2025hi3dgen}, but it is applied in a latent-based pipeline. The DiT prediction in Hi3DGen must first be decoded by a VAE decoder before a mesh can be extracted and rendered for normal supervision. This additional decoding step is feasible at relatively low resolutions such as $256^3$, but becomes memory-intensive at high resolutions such as $1536^3$. In contrast, our latent-free DiT outputs the SDF voxel grid directly, so Marching Cubes and normal rendering can be applied without any VAE decoding. This makes native geometric optimization practical even at high resolution.

\subsection{Camera Parameter Prediction}
\label{sec:camera-prediction}
To strengthen the spatial correspondence between the input image and the generated geometry, we further add a camera prediction branch to the latent-free DiT. Since Direct3D-V3 targets object-level generation rather than scene reconstruction, we adopt an object-centric look-at camera model. In this setting, the object is assumed to be centered in a canonical coordinate frame, and the camera is described by its viewing direction, distance, field of view, and image-plane offset. This avoids unnecessary free variables for object-level conditioning while still capturing the view variation needed to align the generated geometry with the conditioning image.

The branch starts from a learnable camera token $\mathbf{q}_{\mathrm{cam}}$, which first attends to the patch-level DiT output $\mathbf{s}_p$ and then attends to the DINOv2 image tokens $\mathbf{z}$:
\begin{equation}
    \bar{\mathbf{q}}_{\mathrm{cam}}
    =
    \mathrm{CrossAttn}\!\left(
    \mathrm{CrossAttn}\!\left(\mathbf{q}_{\mathrm{cam}}, \mathbf{s}_p\right),
    \mathbf{z}
    \right).
\end{equation}
The updated camera token is then passed through an MLP to predict the following camera vector:
\begin{equation}
    \hat{\boldsymbol{\kappa}}
    =
    \mathrm{MLP}\!\left(\bar{\mathbf{q}}_{\mathrm{cam}}\right),
    \qquad
    \boldsymbol{\kappa}
    =
    (sin(\theta), cos(\theta),\phi,\rho,\psi,c_x,c_y).
\end{equation}
Here $\theta$ and $\phi$ denote azimuth and elevation, $\rho=\log d$ denotes log camera distance, $\psi=\log \tan fov$ denotes log field of view, and $(c_x,c_y)$ are the principal point offsets. All the target parameters are normalized to $[-1,1]$. We supervise this branch with
\begin{equation}
    \mathcal{L}_{\mathrm{cam}}
    =
    \left\|
    \hat{\boldsymbol{\kappa}}
    -
    \boldsymbol{\kappa}^{\mathrm{gt}}
    \right\|_1 .
\end{equation}
This compact parameterization is well suited to single-object generation. The azimuth and elevation provide a normalized representation of view direction, while $\log d$ and $\log \tan fov$ stabilize the dynamic range of distance and field of view. The principal point offsets $(c_x,c_y)$ capture image-plane shifts with only two normalized scalars. As an auxiliary task, camera prediction encourages the DiT to learn the mapping between image observations and 3D space, which improves the alignment between the generated mesh and the conditioning image.

The final training objective combines flow matching, implicit manifold regularization, native geometric optimization, and camera prediction:
\begin{equation}
    \mathcal{L}
    =
    \mathcal{L}_{\mathrm{fm}}
    +
    \lambda_{\mathrm{imr}}\mathcal{L}_{\mathrm{imr}}
    +
    \lambda_{\mathrm{geo}}\mathcal{L}_{\mathrm{geo}}
    +
    \lambda_{\mathrm{cam}}\mathcal{L}_{\mathrm{cam}},
\end{equation}
where $\lambda_{\mathrm{imr}}$, $\lambda_{\mathrm{geo}}$, and $\lambda_{\mathrm{cam}}$ control the strengths of the corresponding auxiliary losses.

\section{Application: Generative Reconstruction}
\label{sec:reconst}

\begin{figure*}[!t]
  \centering
  \includegraphics[width=\linewidth]{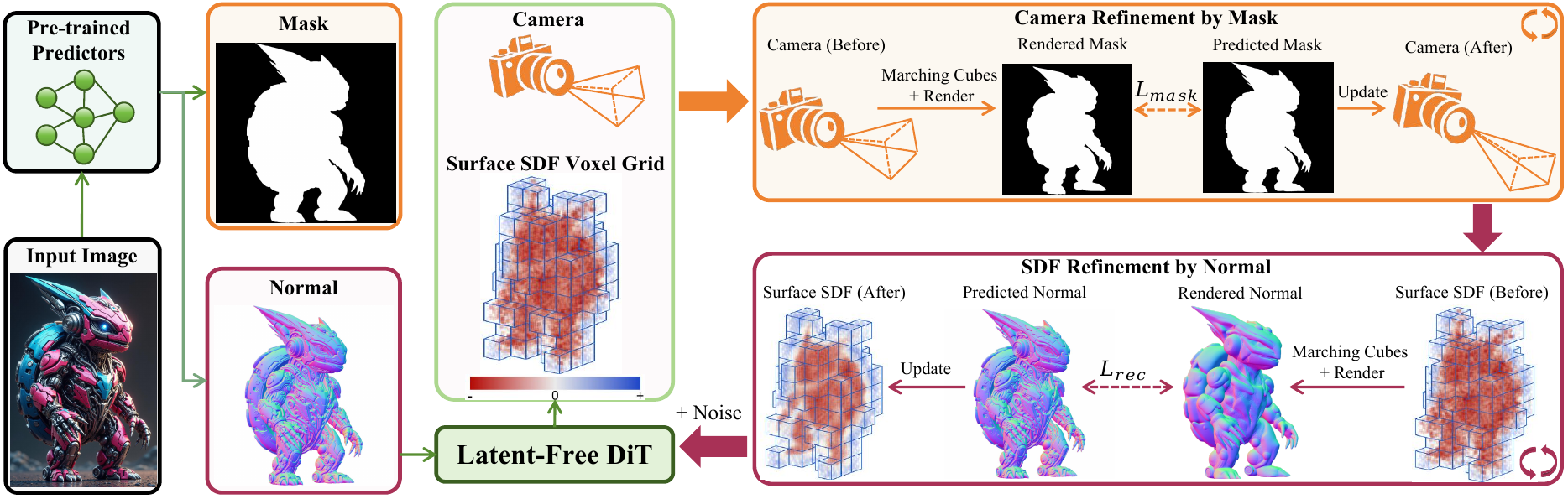}
  \caption{Generative reconstruction pipeline. Given the input images, we use pre-trained predictors to estimate the corresponding foreground mask and normal map as auxiliary observations. The key reconstruction mechanism lies in how these cues are coupled with the latent-free DiT output: the mask calibrates the predicted camera, while the normal map drives direct refinement of the predicted SDF voxel grid.
  At each guided sampling step, the DiT proposes a clean SDF grid together with an image-view camera. We first align the rendered mask with the estimated one to correct the camera, and then compare rendered surface normals with the
  estimated normal map to update the SDF values. The refined SDF is re-noised and passed back to the sampler, coupling observation-driven correction with the learned generative prior for the final reconstruction.}
  \Description{}
  \label{fig:reconst}
\end{figure*}

The latent-free DiT can be used not only as a generator, but also as a native 3D prior for reconstruction. Given a single input image, standard image-to-3D generation samples a plausible complete shape conditioned on the image, but the generated surface may still deviate from the observed view in fine geometric detail. Our framework exposes two quantities that make test-time reconstruction natural: the DiT predicts the camera parameters of the conditioning image, and its clean prediction is the SDF voxel grid itself. We therefore refine the sampling trajectory directly in the SDF domain by comparing normals rendered from the predicted mesh with normals estimated from the input image. This turns the generator into a reconstruction prior: the diffusion trajectory maintains plausible unseen geometry, while the visible surface is corrected by an explicit geometric signal from the input view. The generative reconstruction pipeline is illustrated in Figure~\ref{fig:reconst}.

\subsection{Native Geometry as a Reconstruction Prior}
\label{sec:native-reconstruction-prior}

The central advantage of our latent-free formulation is that every clean DiT prediction can be immediately treated as geometry. Let $\mathbf{I}$ be the input image, $\mathbf{M}_{\mathbf{I}}$ be its foreground mask, and $\Phi_{\mathrm{n}}$ be a pretrained normal prediction model such as Lotus-2~\cite{he2025lotus2}. We estimate an input view normal map $\mathbf{N}_{\mathbf{I}}=\Phi_{\mathrm{n}}(\mathbf{I})$, which provides a dense geometric target over the visible object region. During sampling, the DiT predicts a clean SDF voxel grid $\hat{\mathbf{x}}_0$ together with an initial camera estimate $\hat{\boldsymbol{\kappa}}$:
\begin{equation}
    (\hat{\mathbf{x}}_0,\hat{\boldsymbol{\kappa}})
    =
    f_{\boldsymbol{\theta}}(\mathbf{x}_t,t,\mathbf{I}).
\end{equation}
Because $\hat{\mathbf{x}}_0$ is already a raw SDF grid, we can extract a mesh $\hat{\mathbf{M}}=\mathrm{MC}(\hat{\mathbf{x}}_0)$ and render it under $\hat{\boldsymbol{\kappa}}$ without invoking a VAE decoder.

This native geometry path is what makes the reconstruction procedure practical. In a latent-based pipeline, each inner optimization step must decode an intermediate latent representation into geometry before mesh extraction and rendering, which makes iterative test-time refinement expensive at high resolution. In our framework, Marching Cubes and differentiable rendering are applied directly to the DiT output, and gradients can be propagated back to the SDF grid itself. The normal predictor supplies a view-aligned geometric observation, the camera branch supplies the approximate image-to-shape pose, and the latent-free SDF output supplies the optimizable geometry. Together, these components form a closed reconstruction loop in the native geometry domain.

\subsection{Observation Alignment in the Predicted View}
\label{sec:predicted-camera-alignment}

Before applying normal guidance, we align the predicted view with the input observation. Although the camera branch is trained to predict the relative camera parameters of the conditioning image, small errors in azimuth, elevation, distance, field of view, or image center can shift the rendered normal map relative to the input. Optimizing SDF values under such a misaligned view would push the geometry in the wrong direction. We therefore first refine the predicted camera using the foreground mask, which is robust to local normal ambiguity.

Starting from the DiT prediction $\hat{\boldsymbol{\kappa}}$, we render the geometry mask of the predicted mesh and compare it with the input foreground mask:
\begin{equation}
    \mathcal{L}_{\mathrm{mask}}(\boldsymbol{\kappa})
    =
    \left\|
    \mathcal{R}_{\mathrm{m}}
    \!\left(
    \mathrm{MC}(\hat{\mathbf{x}}_0),
    \boldsymbol{\kappa}
    \right)
    -
    \mathbf{M}_{\mathbf{I}}
    \right\|_2^2 ,
\end{equation}
where $\mathcal{R}_{\mathrm{m}}$ denotes differentiable mask rendering. The corrected camera $\boldsymbol{\kappa}^{\star}$ is obtained by gradient descent on the camera vector, initialized from $\hat{\boldsymbol{\kappa}}$. In practice, the DiT already predicts a reliable pose, so this step is a local calibration rather than a full pose search; about 50 iterations are sufficient for convergence in our setting. Importantly, this calibration modifies only the camera parameters and leaves the generated geometry unchanged.

With the corrected camera $\boldsymbol{\kappa}^{\star}$, the input normal map becomes an aligned geometric target for SDF refinement. Given a current SDF grid $\mathbf{x}$, we extract a mesh by Marching Cubes, render its normal map under $\boldsymbol{\kappa}^{\star}$, and compare it with the predicted input normals inside the foreground mask:
\begin{equation}
    \mathcal{L}_{\mathrm{rec}}(\mathbf{x})
    =
    \left\|
    \mathbf{M}_{\mathbf{I}}
    \odot
    \left(
    \mathcal{R}_{\mathrm{n}}
    \!\left(
    \mathrm{MC}(\mathbf{x}),
    \boldsymbol{\kappa}^{\star}
    \right)
    -
    \mathbf{N}_{\mathbf{I}}
    \right)
    \right\|_1 ,
\end{equation}
where $\mathcal{R}_{\mathrm{n}}$ denotes normal rendering. The foreground mask prevents background pixels from affecting the update, while the normal loss focuses the reconstruction signal on visible surface orientation rather than color or texture. This design directly pushes geometric details in the observed view to agree with the input image.

The SDF grid is updated by backpropagation through mesh extraction and rendering:
\begin{equation}
    \mathbf{x}^{r+1}
    =
    \mathbf{x}^{r}
    -
    \alpha_x
    \nabla_{\mathbf{x}^{r}}
    \mathcal{L}_{\mathrm{rec}}(\mathbf{x}^{r}).
\end{equation}
Because the optimized variable is the raw SDF voxel grid, the update directly changes the zero-level set that will be extracted as the final mesh. This differs from latent refinement, where a gradient step updates an abstract latent code and the geometric effect depends on the decoder. Here, the normal loss, rendered mesh, and optimized SDF all live in the same native geometry space, making the update direct and efficient.

\subsection{Generative Refinement along the Sampling Trajectory}
\label{sec:generative-refinement}

\SetAlgoLongEnd
\begin{algorithm}[!t]
\small
\caption{Generative Reconstruction with Native SDF Refinement}
\label{alg:reconst}
\KwIn{Input image $\mathbf{I}$, foreground mask $\mathbf{M}_{\mathbf{I}}$, normal predictor $\Phi_{\mathrm{n}}$, latent-free DiT $f_{\boldsymbol{\theta}}$, reconstruction threshold $t_1$}
\KwOut{SDF voxel grid $\mathbf{x}_0^{\star}$}
\textbf{Prepare observation cues.} Estimate the visible normal map from the input image: $\mathbf{N}_{\mathbf{I}} \leftarrow \Phi_{\mathrm{n}}(\mathbf{I})$\;
\textbf{Initialize generation.} Draw the initial noisy SDF sampling state $\mathbf{x}_t$ from Gaussian noise\;
\textbf{Initialize reconstruction states.} Set the calibrated camera and current clean SDF to empty: $\boldsymbol{\kappa}^{\star} \leftarrow \varnothing$, $\mathbf{x}_0^{\star} \leftarrow \varnothing$\;
\For{each sampling step $t \rightarrow t_{\mathrm{next}}$}{
    Use the latent-free DiT to estimate the clean SDF grid and the camera of the conditioning image: $(\hat{\mathbf{x}}_0,\hat{\boldsymbol{\kappa}}) \leftarrow f_{\boldsymbol{\theta}}(\mathbf{x}_t,t,\mathbf{I})$\;
    \If{$t < t_1$ and $\boldsymbol{\kappa}^{\star}=\varnothing$}{
        Initialize the camera with the DiT pose prediction: $\boldsymbol{\kappa} \leftarrow \hat{\boldsymbol{\kappa}}$\;
        \For{$r=1$ \KwTo $K_{\mathrm{cam}}$}{
            $\mathcal{L}_{\mathrm{mask}} \leftarrow
            \|\mathcal{R}_{\mathrm{m}}(\mathrm{MC}(\hat{\mathbf{x}}_0),\boldsymbol{\kappa})
            -\mathbf{M}_{\mathbf{I}}\|_2^2$\;
            Update the camera parameters using $\nabla_{\boldsymbol{\kappa}}\mathcal{L}_{\mathrm{mask}}$;
        }
        Store the calibrated camera: $\boldsymbol{\kappa}^{\star} \leftarrow \boldsymbol{\kappa}$\;
    }
    \eIf{$t < t_1$}{
        Start from the current clean SDF prediction: $\mathbf{x}_0^{\star} \leftarrow \hat{\mathbf{x}}_0$\;
        \For{$r=1$ \KwTo $K_{\mathrm{sdf}}$}{
            Render surface normals under the calibrated camera and compare them with the input normal target:
            $\mathcal{L}_{\mathrm{rec}} \leftarrow
            \|\mathbf{M}_{\mathbf{I}}\odot(
            \mathcal{R}_{\mathrm{n}}(\mathrm{MC}(\mathbf{x}_0^{\star}),\boldsymbol{\kappa}^{\star})
            -\mathbf{N}_{\mathbf{I}})\|_1$\;
            Update the SDF voxel values using $\nabla_{\mathbf{x}_0^{\star}}\mathcal{L}_{\mathrm{rec}}$ to move the visible zero-level set toward the observation\;
        }
    }{
        Keep the DiT clean prediction unchanged: $\mathbf{x}_0^{\star} \leftarrow \hat{\mathbf{x}}_0$\;
    }
    Re-noise the selected clean SDF estimate for the next DiT step:
    $\mathbf{x}_{t_{\mathrm{next}}} \leftarrow
    t_{\mathrm{next}} \cdot \mathbf{x}_0^{\star}+(1-t_{\mathrm{next}}) \cdot \boldsymbol{\epsilon}$\;
}
\Return{$\mathbf{x}_0^{\star}$}\;
\end{algorithm}
\SetAlgoShortEnd

We interleave SDF refinement with diffusion sampling rather than applying it only after generation finishes. We apply reconstruction guidance only when $t<t_1$: although the current noisy state $\mathbf{x}_t$ is not yet a final geometry, the DiT already predicts a clean SDF estimate $\hat{\mathbf{x}}_0$, which can be rendered and corrected to steer the subsequent generative trajectory. Once $t\geq t_1$, we stop reconstruction guidance and let the DiT continue standard denoising without additional test-time optimization.

At each timestep, we first determine the clean SDF estimate used for the next sampling update. We set \mbox{$\mathbf{x}_0^{\star}$} to the refined SDF grid when reconstruction guidance is applied, and to the DiT prediction \mbox{$\hat{\mathbf{x}}_0$} otherwise. We then re-noise \mbox{$\mathbf{x}_0^{\star}$} using the same linear schedule as flow matching:
\begin{equation}
    \mathbf{x}_{t_{\mathrm{next}}}
    =
    t_{\mathrm{next}} \cdot \mathbf{x}_0^{\star}
    +
    (1-t_{\mathrm{next}}) \cdot \boldsymbol{\epsilon}.
\end{equation}
This update keeps reconstruction and generation coupled during the guidance window. The normal loss pulls the visible surface toward the input image in the early denoising stage, while the subsequent DiT steps continue to project the sample toward the learned shape distribution. As a result, the method improves alignment with the observed view without turning inference into unconstrained per-example fitting, and the generative prior continues to complete invisible regions and regularize the full shape.

Algorithm~\ref{alg:reconst} summarizes the full procedure. The camera is calibrated once when the sampler first enters the early reconstruction window, because the DiT camera estimate is already close and the mask objective quickly removes the remaining alignment error. SDF refinement is then applied only for steps with $t<t_1$, each time using the corrected camera and the same input normal map. The final output of the algorithm is the last clean SDF voxel grid.

\section{Experiments}
\label{sec:experiments}

\subsection{Datasets and Evaluation Metrics}
\label{sec:exp-setup}

We train Direct3D-V3 on 320K meshes from a high-quality subset of Objaverse-XL~\cite{deitke2023objaversexl}. To examine data scalability, we also conduct an additional training experiment by adding approximately 200K private meshes to this public subset. For each mesh, we render 16 RGB images and 16 normal maps as conditioning images, with randomized viewpoints, illumination, and focal lengths.

For image-to-3D generation, we evaluate on 102 input images from real photographs, web-collected images, and AI-generated images. Since these inputs do not have ground-truth 3D shapes, we measure the consistency between each input image and the generated geometry using two pretrained 3D representation models: Uni3D~\cite{zhou2023uni3d} and ULIP-2~\cite{xue2023ulip2}. These metrics compare the input-image feature with the generated-shape feature, providing complementary estimates of alignment between input images and generated geometry.

For module ablations, we use a held-out Objaverse-XL subset that has no overlap with the training data. To evaluate geometric quality more directly, we report Normal-FID, a normal-rendering variant of the Fr\'echet Inception Distance~\cite{heusel2017gans}. Specifically, for each generated mesh and its corresponding ground-truth mesh, we render normal maps from identical camera views and compute FID between the generated and ground-truth normal renderings. This protocol focuses the ablation study on shape fidelity and surface quality.

For generative reconstruction, we report quantitative results on Google Scanned Objects (GSO)~\cite{downs2022google}. We employ Chamfer Distance (CD), F1-score at 0.95 and 0.99, and Volume IoU to evaluate reconstruction quality. Since GSO contains relatively simple scanned objects and does not fully reflect reconstruction quality on complex geometry, we additionally use images generated by AI and images collected from online communities for qualitative comparison.

\subsection{Implementation Details}
\label{sec:implementation-details}

Our latent-free DiT trains and samples directly on raw SDF voxel grids. We use a truncated SDF with a truncation threshold of $1/64$ and a patch size of $p=8$. To obtain the surface patch indices, we follow the setting of Hunyuan3D 2.1~\cite{hunyuan3d2025hunyuan3d} and train a vecset DiT as the first-stage generator, which quickly produces a coarse mesh for surface patch extraction. The patch-level DiT contains 17 blocks with hidden size 1536 and 16 attention heads, while the voxel-level DiT contains 2 blocks with voxel hidden size $D_v=8$. All self-attention layers in the DiT use 3D shifted-window attention for efficiency. Blocks alternate between regular window attention and the same partition shifted by half a window on the 3D patch grid. We set each window to half of the patch grid resolution along every axis; the shifted partition lets consecutive layers cover offset neighborhoods and exchange information across window boundaries without global attention. Image conditioning features are extracted by DINOv2-Large from input images at resolution $1022$. We train the model on 16 NVIDIA A800 GPUs with AdamW and bfloat16 mixed precision, and reduce memory use with gradient checkpointing and FSDP2. The loss weights are set to $\lambda_{\mathrm{imr}}=1\times10^{-5}$, $\lambda_{\mathrm{geo}}=1.0$, and $\lambda_{\mathrm{cam}}=5\times10^{-2}$. Normal maps for geometric supervision are rendered at resolution $1024$. During sampling, we use 15 flow-matching steps and classifier-free guidance scale 4.0. To stabilize high-resolution latent-free training, we adopt a progressive schedule from $256^3$ to $1536^3$ resolution, as summarized in Table~\ref{tab:progressive-training}.

\begin{table}[t]
  \caption{Progressive training schedule for Direct3D-V3.}
  \label{tab:progressive-training}
  \centering
  \begin{tabular}{lccc}
    \toprule
    Resolution & Learning rate & Batch size & Training Steps \\
    \midrule
    $256^3$  & 1e-4 & 8$\times$16 & 400K \\
    $512^3$  & 1e-4 & 4$\times$16 & 50K \\
    $1024^3$ & 5e-5 & 1$\times$16 & 50K \\
    $1536^3$ & 5e-5 & 1$\times$16 & 10K \\
    \bottomrule
  \end{tabular}
\end{table}

For generative reconstruction, we use the same DiT checkpoint as a shape prior and refine the clean SDF prediction during sampling. The mask-based camera calibration uses $K_{\mathrm{cam}}=50$ optimization steps. SDF refinement is applied when $t<t_1$, where $t_1=0.8$, using $K_{\mathrm{sdf}}=100$ gradient steps. We use Lotus-2 as the normal predictor and render normals at resolution $1024$ for the reconstruction loss.

\subsection{Comparison with Image-to-3D Generation Methods}
\label{sec:image-to-3d-comparison}

\begin{figure}[!t]
  \centering
  \includegraphics[width=0.99\linewidth]{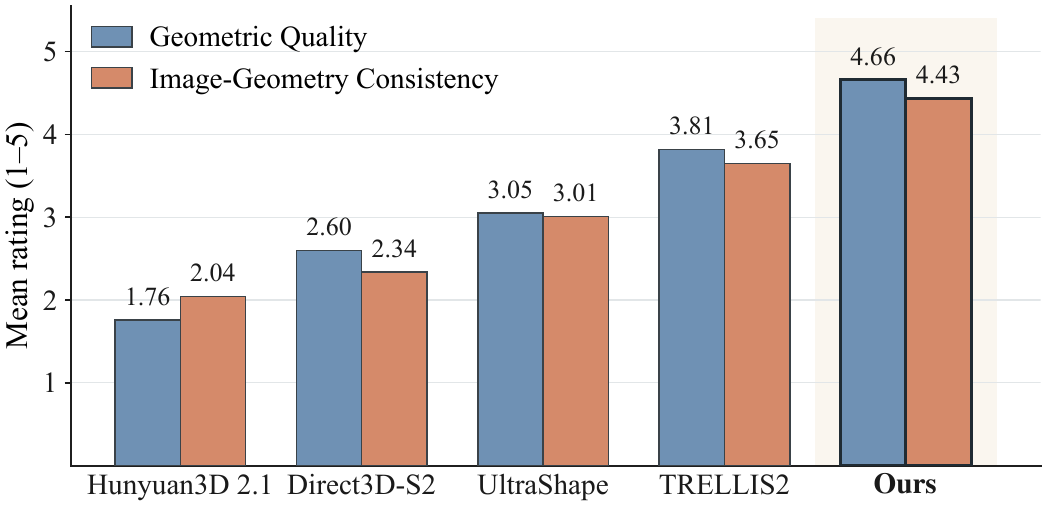}
  \caption{User study results for image-to-3D generation. Scores range from 1 to 5, and higher is better.}
  \Description{}
  \label{fig:user_study}
\end{figure}

We evaluate Direct3D-V3 on the test images described in Section~\ref{sec:exp-setup}, comparing against recent open-source image-to-3D methods including Michelangelo~\cite{zhao2023michelangelo}, Hunyuan3D 2.1~\cite{hunyuan3d2025hunyuan3d}, Direct3D-S2~\cite{wu2025direct3ds2}, UltraShape~\cite{jia2025ultrashape}, and TRELLIS2~\cite{xiang2026native}. Table~\ref{tab:image-to-3d-comparison} reports quantitative results using Uni3D and ULIP-2, which measure consistency between the input image and generated shape in pretrained 3D feature spaces. Direct3D-V3 outperforms all competing methods on both metrics, and adding private training data further improves the scores. However, Uni3D and ULIP-2 mainly reflect semantic alignment with the conditioning image and cannot directly measure geometric precision, such as surface regularity, thin structures, and local details. We therefore conduct a user study with 1--5 quality ratings. As shown in Figure~\ref{fig:user_study}, Direct3D-V3 receives the highest user preference, indicating that users favor the geometric quality of our generated meshes.

\begin{table}[t]
  \caption{Quantitative comparison with recent image-to-3D methods. Higher is better for all metrics, and the best result is shown in bold. Ours$^*$ denotes Direct3D-V3 trained with the public training data plus approximately 200K additional private meshes.}
  \label{tab:image-to-3d-comparison}
  \centering
  \setlength{\tabcolsep}{15pt}
  \begin{tabular}{@{}lcc@{}}
    \toprule
    Method & Uni3D $\uparrow$ & ULIP-2 $\uparrow$ \\
    \midrule
    Michelangelo & 0.3404 & 0.3605 \\
    Hunyuan3D 2.1 & 0.4187 & 0.4424 \\
    Direct3D-S2 & 0.4190 & 0.4485 \\
    UltraShape & 0.4185 & 0.4510 \\
    TRELLIS2 & 0.4172 & 0.4405 \\
    Ours & 0.4198 & 0.4535 \\
    Ours$^*$ & \textbf{0.4224} & \textbf{0.4547} \\
    \bottomrule
  \end{tabular}
\end{table}

\begin{figure*}[!ht]
  \centering
  \includegraphics[width=0.93\linewidth]{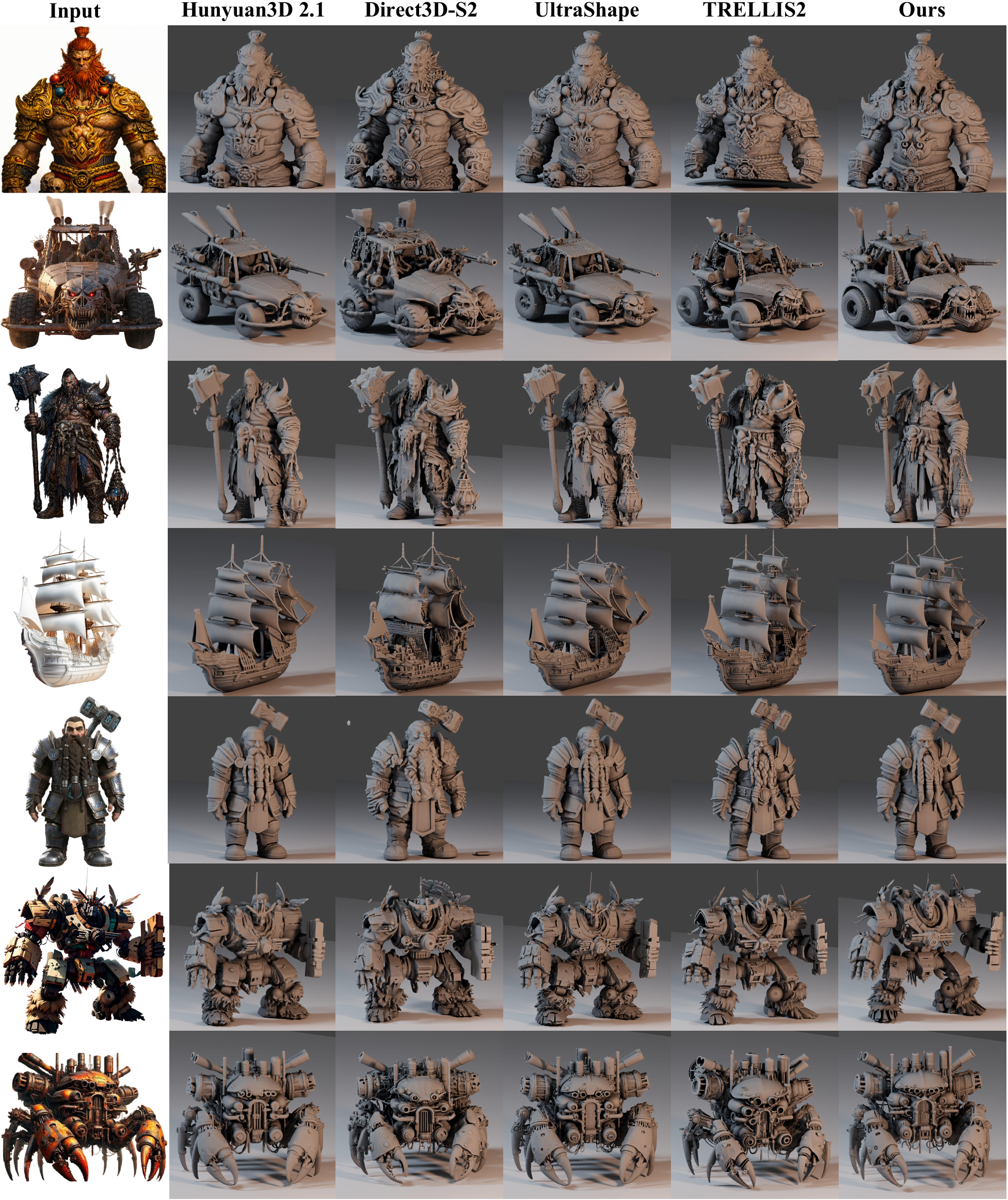}
  \caption{Qualitative comparison with recent open-source image-to-3D generation methods.}
  \Description{}
  \label{fig:compare_gen}
\end{figure*}

Figure~\ref{fig:compare_gen} provides a qualitative comparison of generated meshes. Hunyuan3D 2.1 usually captures the coarse object shape and produces stable overall geometry, but it tends to miss or smooth out fine structures, especially on hair, beard, ornaments, and other high-frequency regions. Direct3D-S2 generates richer local details, yet these details are sometimes less organized: surface patterns can become cluttered, and the generated geometry may drift from the conditioning image when the input contains dense structures or ambiguous contours. UltraShape follows the input silhouette well, but its geometric precision still has a visible ceiling on thin parts and high-frequency surface patterns. TRELLIS2 is the closest high-resolution competitor, since both TRELLIS2 and Direct3D-V3 scale to $1536^3$ generation and can recover substantially more detail than lower-resolution methods. Even so, Direct3D-V3 produces more faithful and regular local geometry in several representative cases: it better preserves hair and beard structures in the first and fifth examples, sharper tire tread patterns in the second example, more complete facial details in the third example, and more regular ropes on the ship in the fourth example. Moreover, unlike TRELLIS2 and Direct3D-S2, our approach is implemented as a pure transformer architecture and does not require specialized sparse 3D convolution infrastructure, making the high-resolution generator simpler while remaining competitive with the strongest sparse 3D frameworks.

\subsection{Generative Reconstruction Results}
\label{sec:generative-reconstruction-results}

We further evaluate our Direct3D-V3 in the single-view generative reconstruction setting. We compare against TRELLIS~\cite{xiang2025structured}, TRELLIS2~\cite{xiang2026native}, CUPID~\cite{huang2025cupid}, and ReconViaGen~\cite{chang2025reconviagen}. Given a single image, Direct3D-V3 first samples a complete shape from the learned prior, uses the predicted camera parameters to establish the input view, and refines the native SDF prediction with normal-guided test-time optimization. Since the refinement is applied directly to the SDF grid, the observed surface can be corrected in the geometry domain while the diffusion prior continues to regularize unobserved regions.

\begin{table}[t]
  \caption{Quantitative comparison on GSO generative reconstruction. The table reports Chamfer Distance (CD), F1-score at 0.95 and 0.99, and Volume IoU. The best result is shown in bold.}
  \label{tab:generative-reconstruction-comparison}
  \centering
  \setlength{\tabcolsep}{3pt}
  \begin{tabular}{@{}lcccc@{}}
    \toprule
    Method & CD $\downarrow$ & F1$_{0.95}$ $\uparrow$ & F1$_{0.99}$ $\uparrow$ & Vol. IoU $\uparrow$ \\
    \midrule
    TRELLIS & 0.0153 & 0.9586 & 0.5123 & 0.3315 \\
    CUPID & 0.0141 & 0.9550 & 0.5728 & 0.3744 \\
    ReconViaGen & 0.0194 & 0.9239 & 0.4585 & 0.2830 \\
    TRELLIS2 & 0.0147 & 0.9510 & 0.5934 & \textbf{0.3884} \\
    Ours & \textbf{0.0131} & \textbf{0.9600} & \textbf{0.6043} & 0.3815 \\
    \bottomrule
  \end{tabular}
\end{table}

\begin{figure*}[!t]
  \centering
  \includegraphics[width=0.98\linewidth]{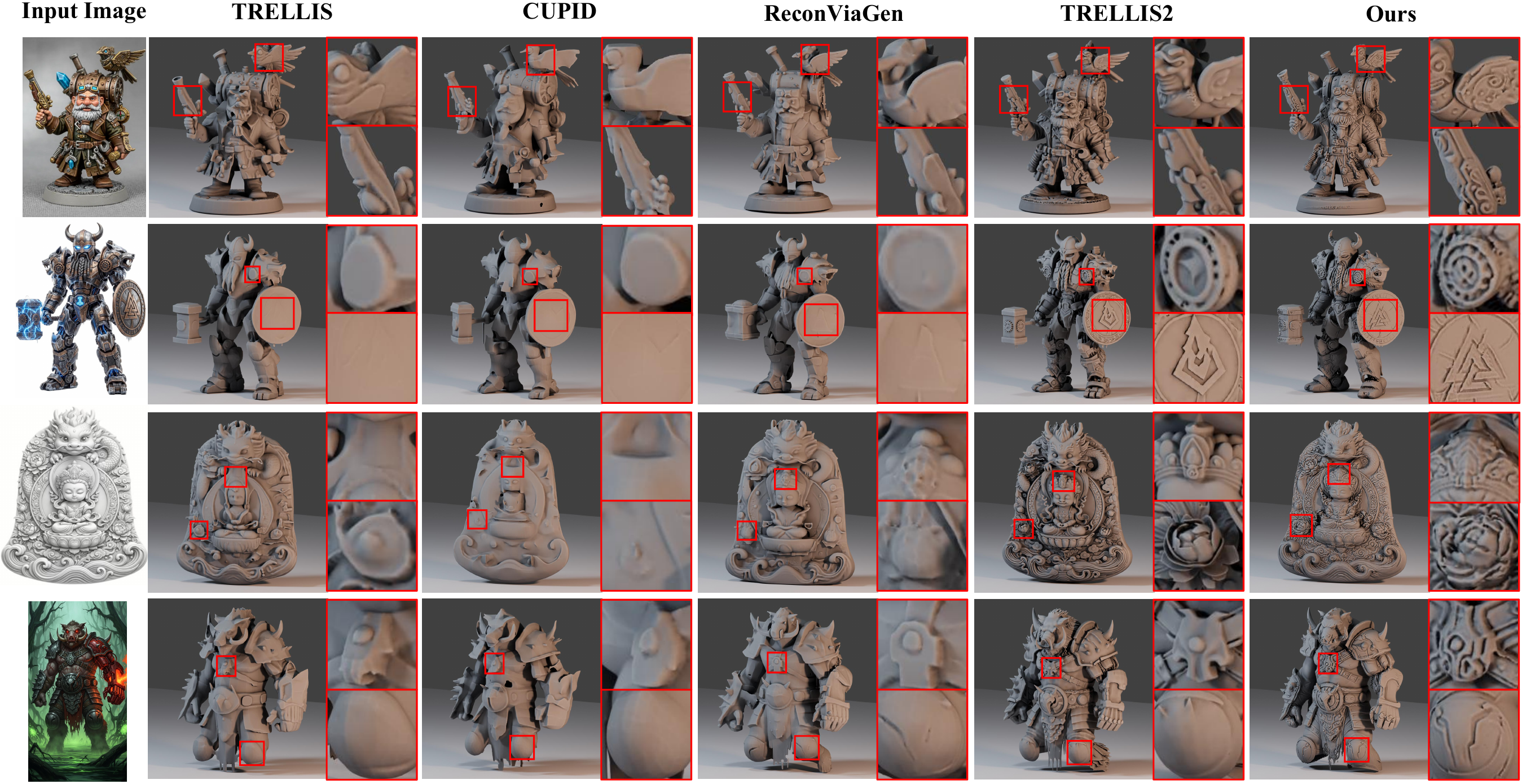}
  \caption{Qualitative comparison for single-view generative reconstruction. The figure compares reconstructed meshes from the same challenging input images.}
  \Description{}
  \label{fig:compare_reconst}
\end{figure*}

Table~\ref{tab:generative-reconstruction-comparison} reports reconstruction accuracy on GSO using CD, F1-score at 0.95 and 0.99, and Volume IoU. Direct3D-V3 achieves the best CD and F1 scores among all compared methods, while obtaining a competitive Volume IoU slightly below TRELLIS2. Although these results validate our reconstruction framework on a standard benchmark, GSO contains relatively simple scanned objects and does not fully reveal how different methods handle complex geometry. We therefore further compare on AI images and community images, which contain denser local structures, richer visual styles, and more challenging fine detail. As shown in Figure~\ref{fig:compare_reconst}, CUPID, ReconViaGen, TRELLIS, and TRELLIS2 recover plausible complete shapes, but their surfaces are often over-smoothed or locally inconsistent with the input image on thin structures, concave parts, and fine details. Direct3D-V3 reconstructs more faithful visible surface geometry, with stronger alignment to the input image, sharper local details, and coherent global completion.

\begin{figure}[!t]
  \centering
  \includegraphics[width=0.98\linewidth]{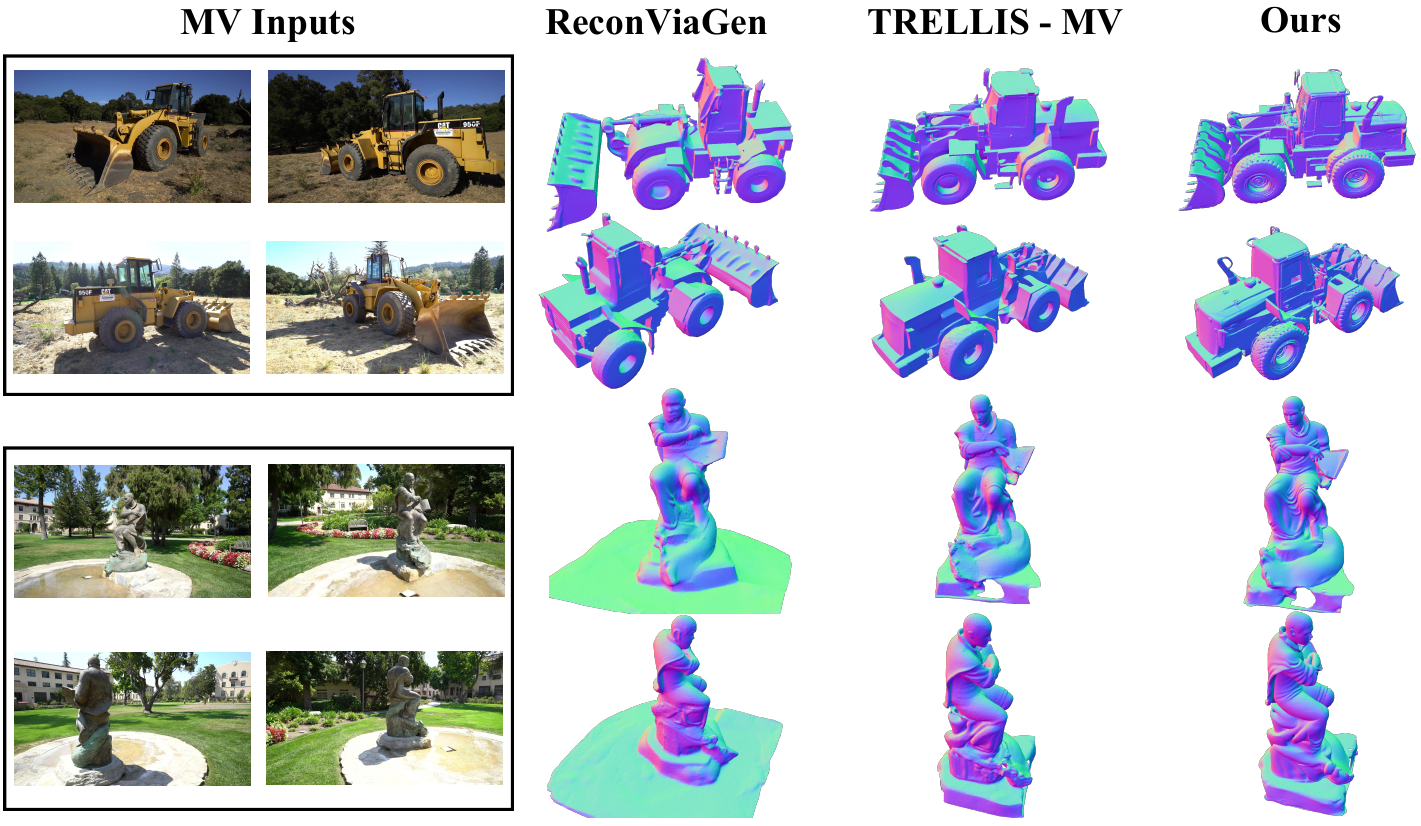}
  \caption{Qualitative comparison for multi-view generative reconstruction on scenes from the Tanks and Temples dataset.}
  \Description{}
  \label{fig:compare_mv_reconst}
\end{figure}

We also extend our framework to multi-view inputs with minimal changes. We concatenate the DINOv2 image tokens extracted from all views and inject them into the latent-free DiT through the same cross-attention layers, fine-tuning the model for 20K steps. At inference time, we use the mesh generated by the TRELLIS MultiDiffusion pipeline to obtain the surface patch indices. We evaluate this setting on scenes from the standard Tanks and Temples dataset~\cite{knapitsch2017tanks}; the qualitative results are shown in Figure~\ref{fig:compare_mv_reconst}, where our method reconstructs meshes with sharper and more complete surface details than competing approaches.

\subsection{Ablation Study}
\label{sec:ablation-study}

All ablations are conducted on the held-out Objaverse-XL subset described in Section~\ref{sec:exp-setup}. By default, we extract surface patch indices directly from the ground-truth meshes to isolate the ablation results from errors introduced by the first-stage generator. We report Normal-FID computed from rendered normal maps, following the evaluation protocol in Section~\ref{sec:exp-setup}. Unless otherwise specified, we run ablation studies at $256^3$ resolution for efficiency.

\begin{table}[t]
    \caption{Core component ablations on the held-out Objaverse-XL subset. Each checkmark indicates that the corresponding module is enabled. Lower is better for Normal-FID, and the
  best result is shown in bold.}
    \label{tab:ablation-core-components}
    \centering
    \begin{tabular}{@{}ccccc@{}}
      \toprule
      Patch DiT & Voxel DiT & IMR & Normal Sup. & Normal-FID $\downarrow$ \\
      \midrule
      $\checkmark$ & - & - & - & 92.96 \\
      $\checkmark$ & $\checkmark$ & - & - & 75.80 \\
      $\checkmark$ & $\checkmark$ & $\checkmark$ & - & 73.24 \\
      $\checkmark$ & $\checkmark$ & $\checkmark$ & $\checkmark$ & \textbf{68.97} \\
      \bottomrule
    \end{tabular}
  \end{table}

\begin{table}[t]
  \caption{Effect of IMR on boundary artifacts. Lower boundary-edge count indicates fewer open mesh boundaries.}
  \label{tab:imr-boundary-edges}
  \centering
  \begin{tabular}{@{}ccc@{}}
    \toprule
    Resolution & Method & Avg. boundary-edge count $\downarrow$ \\
    \midrule
    \multirow{2}{*}{$256^3$}  & w/o IMR & 2,462 \\
                                   & w/ IMR  & \textbf{196} \\
    \midrule
    \multirow{2}{*}{$1024^3$} & w/o IMR & 25,353 \\
                                   & w/ IMR  & \textbf{898} \\
    \bottomrule
  \end{tabular}
\end{table}

\begin{figure}[!t]
  \centering
  \includegraphics[width=1.0\linewidth]{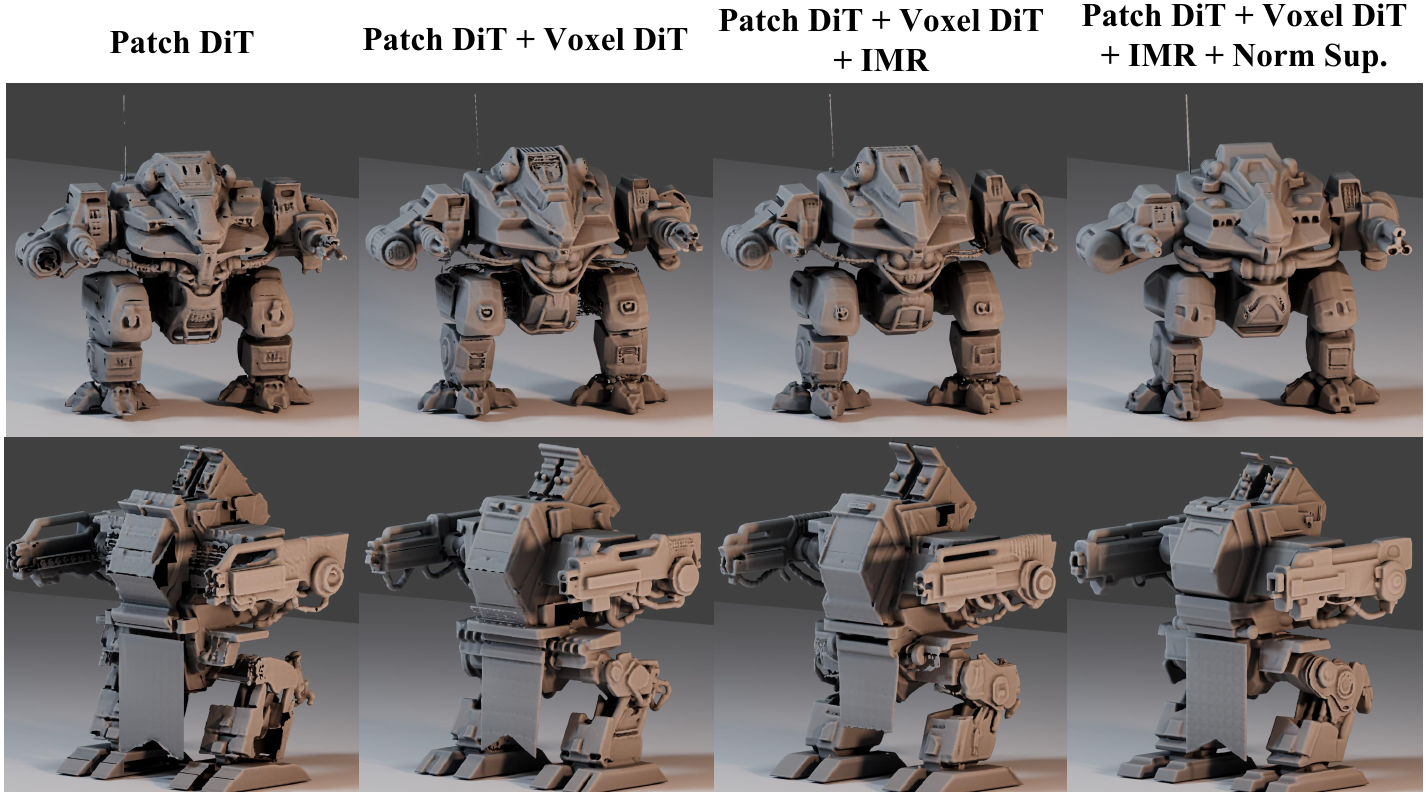}
  \caption{Qualitative ablation of the core components. The patch-level DiT alone produces coarse and incomplete geometry. Adding the voxel-level DiT restores finer details but still leaves many holes; IMR greatly reduces these artifacts, and normal-rendering supervision further improves the quality.}
  \Description{Qualitative comparison of core component ablations, showing improvements from patch-level DiT to voxel-level DiT, IMR, and normal-rendering supervision.}
  \Description{}
  \label{fig:ablation-core}
\end{figure}

\textbf{Effect of each core component.} Table~\ref{tab:ablation-core-components} and Figure~\ref{fig:ablation-core} isolate the contribution of each component in the latent-free geometry pipeline. With only the patch-level DiT, the SDF values within each selected patch are encoded as channels of a patch token, so local intra-patch variation is not modeled explicitly at voxel granularity. Adding the voxel-level DiT introduces an intra-patch refinement stage that models the local SDF field directly over voxel tokens, improving surface detail and reducing Normal-FID from 92.96 to 75.80. However, the generated meshes still contain many holes, reflecting the missing manifold prior of a VAE decoder. IMR addresses this issue by regularizing local SDF transitions around the zero level set, greatly reducing holes and improving mesh validity. Table~\ref{tab:imr-boundary-edges} further shows that IMR sharply reduces boundary edges at both \mbox{$256^3$} and \mbox{$1024^3$} resolutions. After this boundary reduction, standard post-processing hole filling can close the remaining small openings and produce watertight meshes. With normal-rendering supervision, the prediction is optimized entirely in the geometry domain after mesh extraction, further improving surface quality and achieving the best Normal-FID of 68.97.

\begin{figure}[!t]
  \centering
  \includegraphics[width=1.0\linewidth]{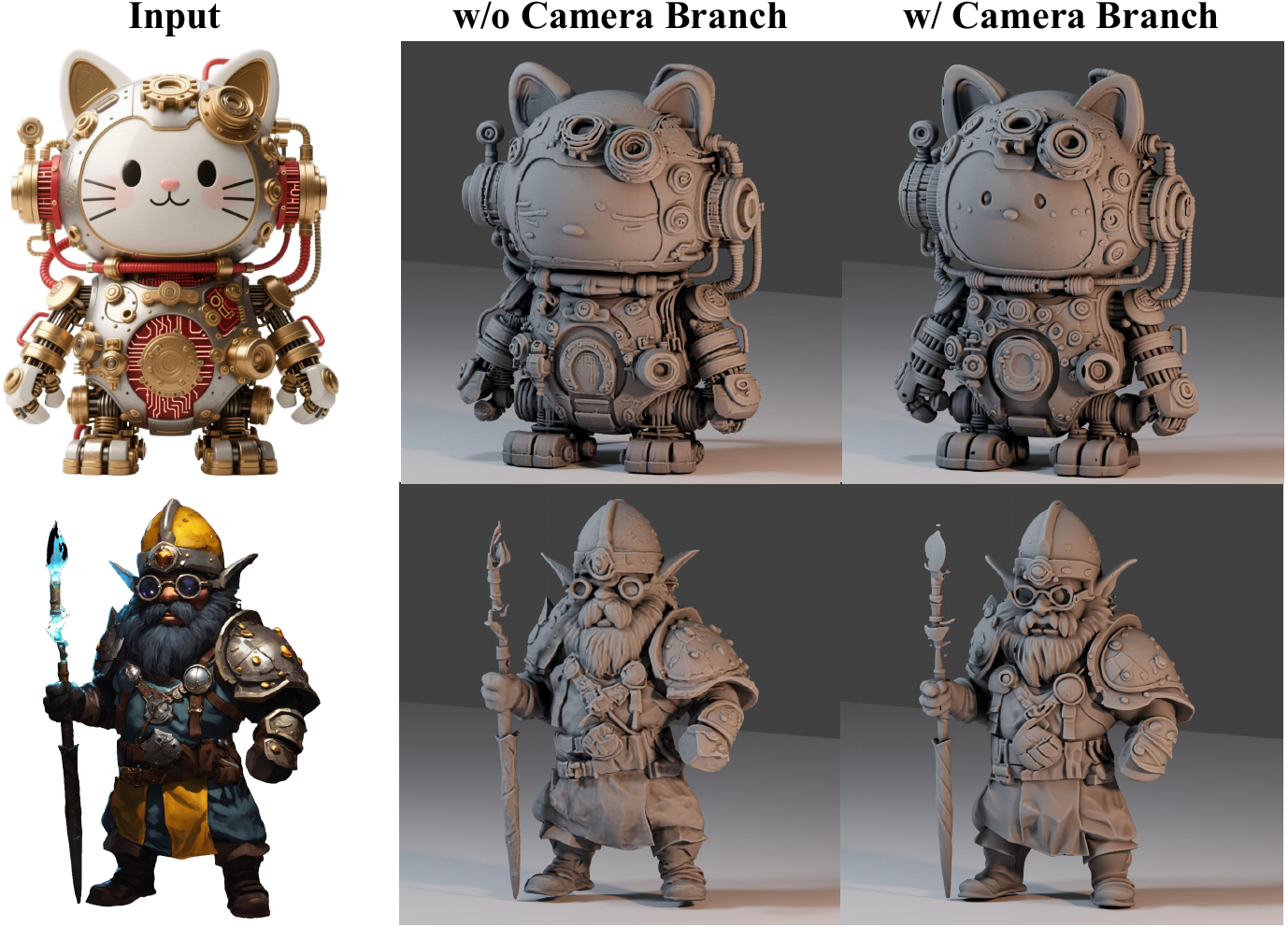}
  \caption{Qualitative ablation of the camera prediction branch. Adding the branch improves alignment between the generated mesh and the conditioning image, especially for structures visible from the conditioning view.}
  \Description{Qualitative comparison showing that the camera prediction branch improves alignment between the conditioning image and generated mesh.}
  \Description{}
  \label{fig:camera}
\end{figure}

\textbf{Effect of camera prediction branch.} Figure~\ref{fig:camera} evaluates the auxiliary camera prediction branch. Without this branch, the generated mesh can still be plausible in overall shape, but the geometry visible from the conditioning view may drift from the input image, leading to mismatched local structures. Adding camera prediction provides an explicit alignment signal during training, encouraging the DiT to associate image observations with the corresponding 3D viewpoint. As a result, the generated mesh aligns better with the input image while preserving the same overall geometry quality.

\begin{table}[t]
  \caption{Patch-size ablation for the latent-free DiT. Smaller patches improve Normal-FID but increase the number of patch tokens; we use $p=8$ as the default patch size.}
  \label{tab:ablation-patch-size}
  \centering
  \setlength{\tabcolsep}{12pt}
  \begin{tabular}{@{}lc@{}}
    \toprule
    Patch size & Normal-FID $\downarrow$ \\
    \midrule
    $p=4$ & \textbf{57.63} \\
    $p=8$ (default) & 68.97 \\
    $p=12$ & 76.21 \\
    \bottomrule
  \end{tabular}
\end{table}

\textbf{Effect of patch size.} Table~\ref{tab:ablation-patch-size} reports the generation quality under different patch sizes. Smaller patches assign each token to a finer SDF region, which preserves more geometric detail and improves Normal-FID. However, reducing the patch size also increases the patch sequence length, leading to higher memory and attention cost. Larger patches are cheaper but blur local structures because each token must represent a broader surface neighborhood. We therefore choose $p=8$ as the default setting, which provides a practical trade-off between geometric fidelity and computational efficiency.

\begin{table}[t]
  \caption{First-stage choices evaluated on the Objaverse-XL validation set at $512^3$ resolution. Lower is better for Normal-FID.}
  \label{tab:first-stage-choices}
  \centering
  \setlength{\tabcolsep}{10pt}
  \begin{tabular}{@{}lc@{}}
    \toprule
    First stage & Normal-FID $\downarrow$ \\
    \midrule
    Dense latent-free DiT & 80.61 \\
    Hunyuan3D 2.1 & 80.43 \\
    Ours (VecSet DiT) & \textbf{76.62} \\
    \bottomrule
  \end{tabular}
\end{table}

\begin{figure}[!t]
  \centering
  \includegraphics[width=1.0\linewidth]{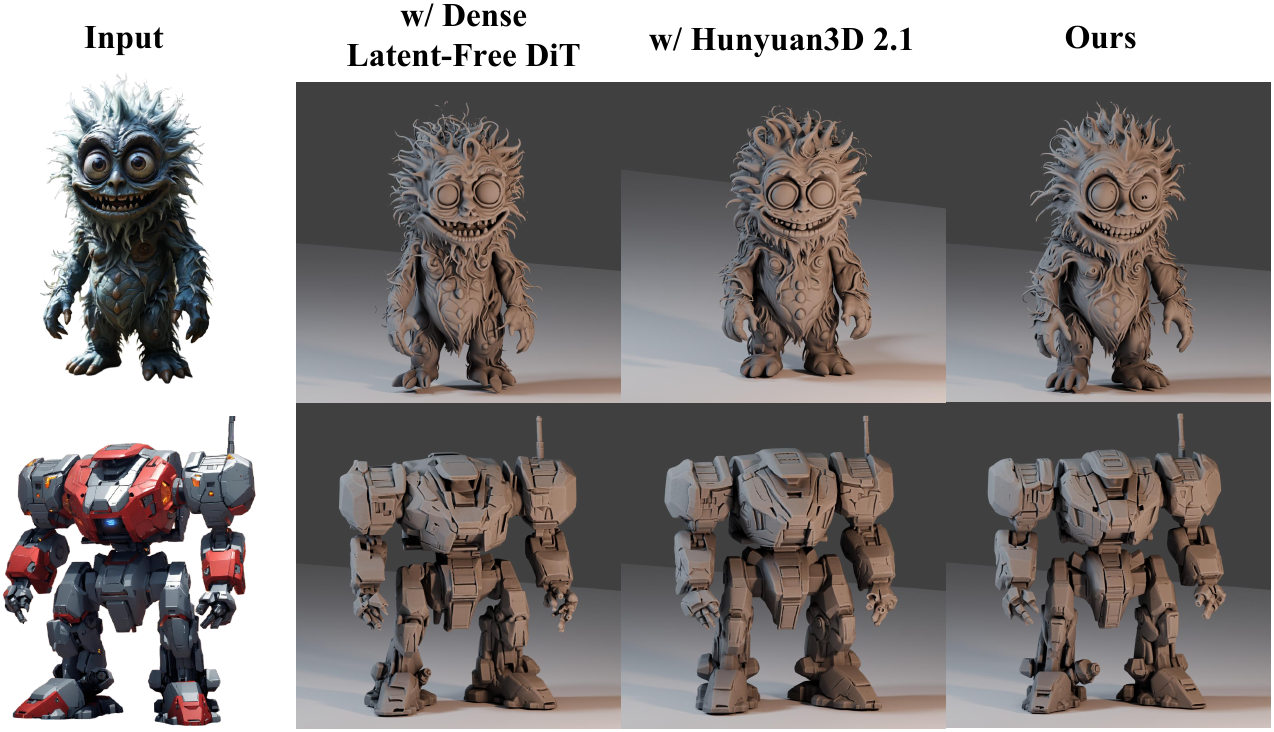}
  \caption{Ablation of first-stage surface patch indices at $512^3$ resolution. Comparable results across a pre-trained VecSet DiT, a dense $64^3$ index predictor, and Hunyuan3D 2.1 indicate that the latent-free DiT is robust to different sources of surface patch indices.}
  \Description{}
  \label{fig:first_stage}
\end{figure}

\textbf{Effect of first-stage surface patch indices.} Figure~\ref{fig:first_stage} compares three sources of surface patch indices at $512^3$ resolution. These indices define the sparse support used by the latent-free DiT. In our default configuration, a pre-trained VecSet DiT generates a coarse mesh, from which we extract the indices. We also train a dense DiT with the same architecture as the latent-free DiT to directly predict a $64^3$ dense voxel grid, where each voxel indicates whether the corresponding region belongs to the surface patch set. This direct predictor performs similarly to our default mesh-based source, showing that the second-stage latent-free DiT is not sensitive to a specific index generator as long as the predicted surface support is reasonable. The VecSet DiT remains preferable for its flexibility: since it outputs a mesh before patch extraction, surface patch indices can be produced at arbitrary resolutions. We further replace the first-stage generator with Hunyuan3D 2.1~\cite{hunyuan3d2025hunyuan3d} and obtain similar results, which further supports the robustness of the latent-free DiT to different sources of surface patch indices.
Table~\ref{tab:first-stage-choices} reports Normal-FID on the Objaverse-XL validation set when the surface patch indices are extracted from different first-stage generators instead of the ground-truth meshes. Among the tested choices, using our VecSet DiT as the first-stage generator achieves the best result.

\section{Conclusion and Limitations}
\label{sec:conclusion}

We presented Direct3D-V3, a latent-free DiT framework that revisits the need for VAE-based compression in high-fidelity 3D generation. Instead of modeling compressed 3D latent representations, Direct3D-V3 performs diffusion in the native voxel domain by directly denoising raw SDF voxel grids. To make this formulation scalable, the model uses sparse surface patch modeling and a dual-granularity DiT, where a patch-level transformer captures global shape structure and a lightweight voxel-level transformer restores local SDF variation. Beyond scalability, Direct3D-V3 makes the native SDF output geometrically reliable by constraining neighboring SDF values with implicit manifold regularization, supervising the extracted surface through rendered normals, and estimating the conditioning camera with an auxiliary branch for better view alignment. The same native SDF voxel output also enables generative reconstruction, where test-time normal guidance refines the observed surface while the learned prior completes unseen regions. Experiments show competitive or superior performance against recent image-to-3D methods, together with finer local geometry.

\begin{figure}[!t]
  \centering
  \includegraphics[width=0.82\linewidth]{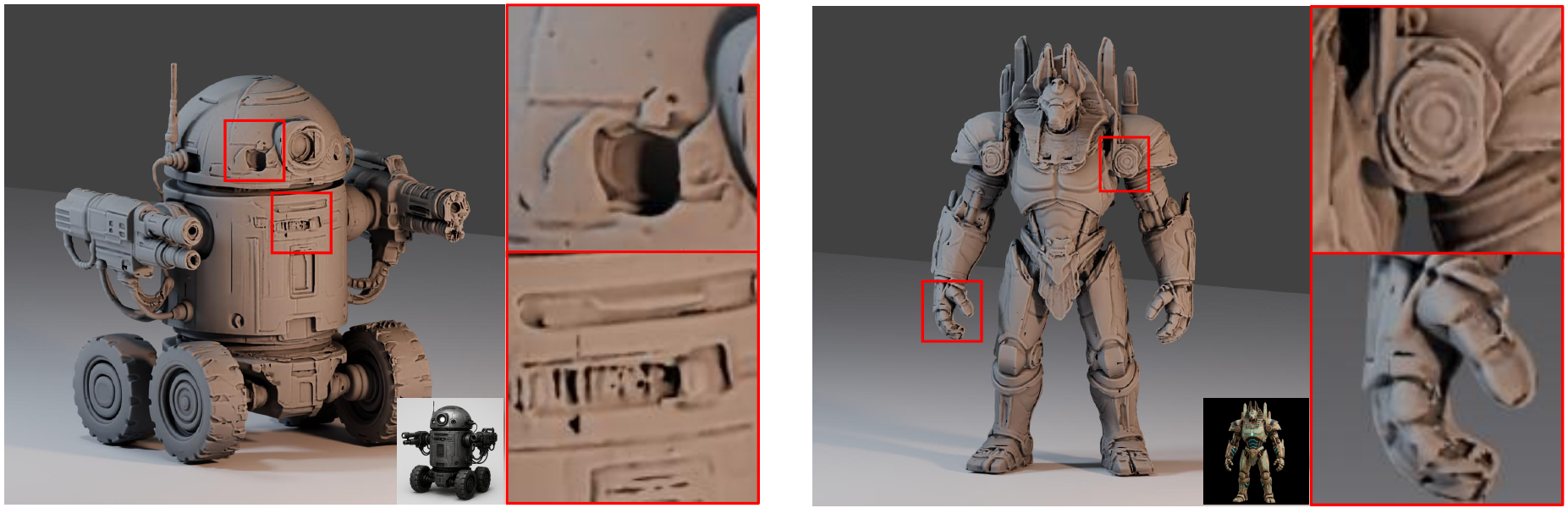}
  \caption{Failure case of single-view generative reconstruction. When the first-stage surface indices are poorly aligned with the conditioning image, normal-guided SDF refinement operates on inaccurate sparse support and can distort the mesh.}
  \Description{A failure case visualization showing distorted geometry caused by normal-guided SDF refinement on inaccurate first-stage surface indices.}
  \label{fig:failure}
\end{figure}

\textbf{Limitations.} The current framework has two main limitations. First, generative reconstruction depends on accurate first-stage surface indices, which define the sparse SDF support used by the latent-free DiT. If this support misses image-observed structures or places patches in inconsistent regions, test-time normal guidance may distort the mesh, especially around thin structures or ambiguous silhouettes. Figure~\ref{fig:failure} visualizes this failure mode, where inaccurate first-stage surface indices cause normal-guided refinement to over-update the native SDF and produce distorted geometry. Improving robustness to inaccurate sparse support remains an important direction, potentially through stronger camera estimation, guidance with uncertainty estimates, or more constrained geometry refinement. Second, Direct3D-V3 currently focuses on geometry only. In principle, the latent-free formulation can be extended beyond SDF values by adding channels for texture, material, or PBR attributes to the voxel grid. However, jointly modeling geometry and appearance in this native space will require additional design to keep training stable and to preserve consistency between surface structure and material signals. We leave this unified latent-free geometry and appearance generation setting for future work.

\begin{acks}
This work was supported by National Key R\&D Program of China (2023YFB3209702), National Natural Science Foundation of China (62472213), Gusu Innovation \& Entrepreneurship Leading Talents Program (ZXL2024361), and computing resources from DreamTech.
\end{acks}

\bibliographystyle{ACM-Reference-Format}
\bibliography{sample-bibliography}

\appendix

\section{Computational Cost and Scaling}
\label{sec:appendix-cost-scaling}

\begin{table}[H]
  \caption{Computational cost and scaling across resolutions and patch sizes. Memory and per-step inference time are measured on one A800 GPU.}
  \label{tab:computational-cost-scaling}
  \centering
  \setlength{\tabcolsep}{3.2pt}
  \begin{tabular}{@{}ccccc@{}}
    \toprule
    Res. & Patch & Avg. Patch Tokens & Memory (MB) & Time/step (s) \\
    \midrule
    \multirow{3}{*}{$256^3$}  & 4  & 7,980   & 4,386  & 0.31 \\
                                   & 8  & 1,629   & 3,716  & 0.12 \\
                                   & 12 & 721     & 3,798  & 0.12 \\
    \midrule
    \multirow{3}{*}{$512^3$}  & 4  & 35,576  & 8,082  & 1.55 \\
                                   & 8  & 7,340   & 5,716  & 0.37 \\
                                   & 12 & 2,986   & 5,080  & 0.25 \\
    \midrule
    \multirow{3}{*}{$1024^3$} & 4  & 153,606 & 21,490 & 12.72 \\
                                   & 8  & 32,536  & 13,312 & 1.82 \\
                                   & 12 & 13,422  & 10,620 & 1.09 \\
    \midrule
    \multirow{3}{*}{$1536^3$} & 4  & 353,637 & 45,128 & 53.76 \\
                                   & 8  & 76,435  & 24,174 & 4.64 \\
                                   & 12 & 31,473  & 19,450 & 2.59 \\
    \bottomrule
  \end{tabular}
\end{table}

\begin{figure*}[!ht]
  \centering
  \includegraphics[width=0.93\linewidth]{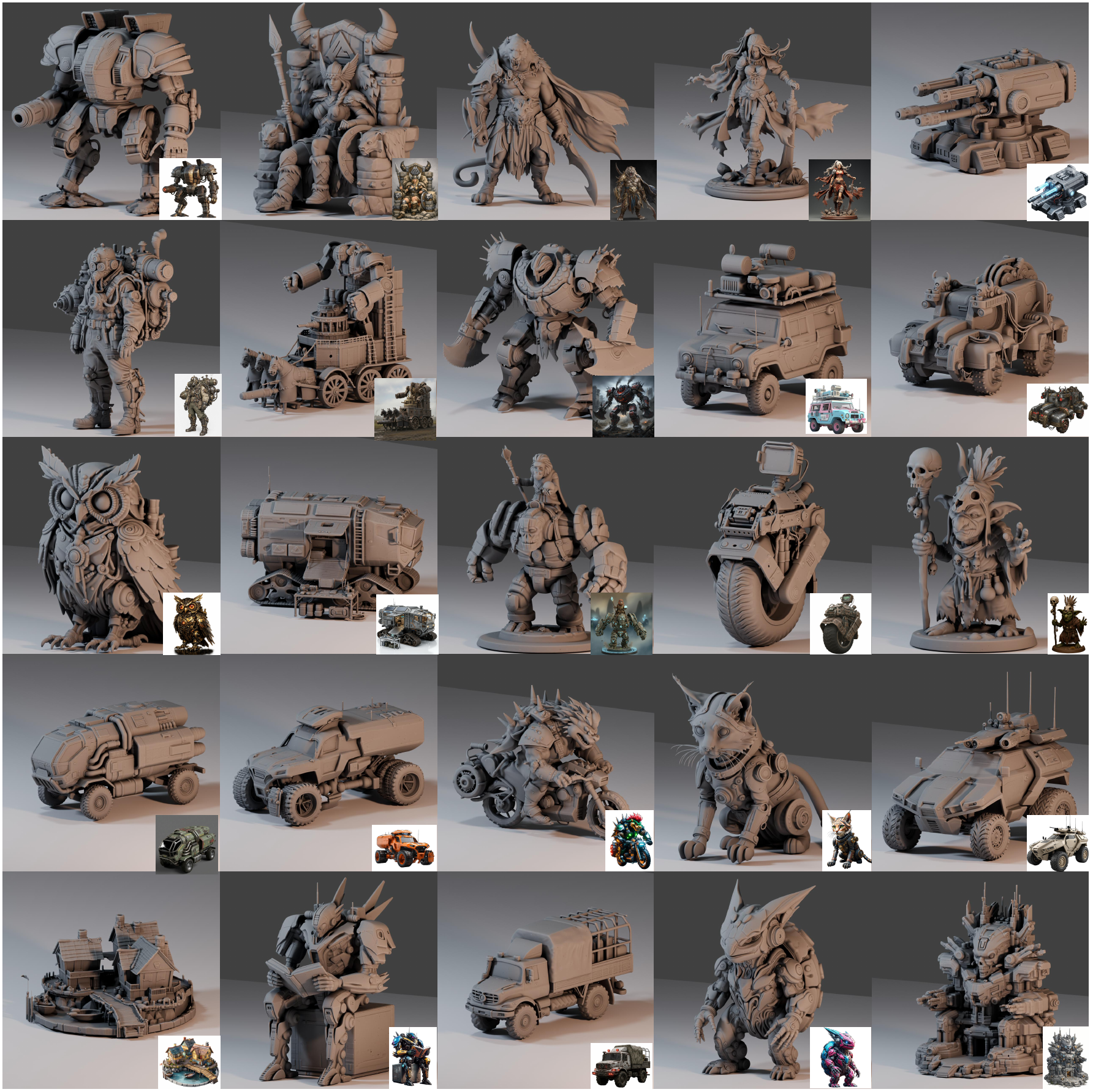}
  \caption{Additional visual results generated by Direct3D-V3.}
  \Description{Additional generated meshes from Direct3D-V3, illustrating diverse shapes and surface details.}
  \label{fig:more_vis}
\end{figure*}

Table~\ref{tab:computational-cost-scaling} reports the average number of active patch tokens, average GPU memory usage, and per-step inference time across resolutions and patch sizes. The cost increases with the number of active patch tokens, making smaller patches more expensive at high resolution. We use patch size 8 as a practical balance between geometric fidelity and efficiency, which supports native SDF generation up to \mbox{$1536^3$}.

\section{Additional Visual Results}

Figure~\ref{fig:more_vis} shows more visual results generated by Direct3D-V3. 

\section{Additional Qualitative Comparisons}

\begin{figure*}[!ht]
  \centering
  \includegraphics[width=0.9\linewidth]{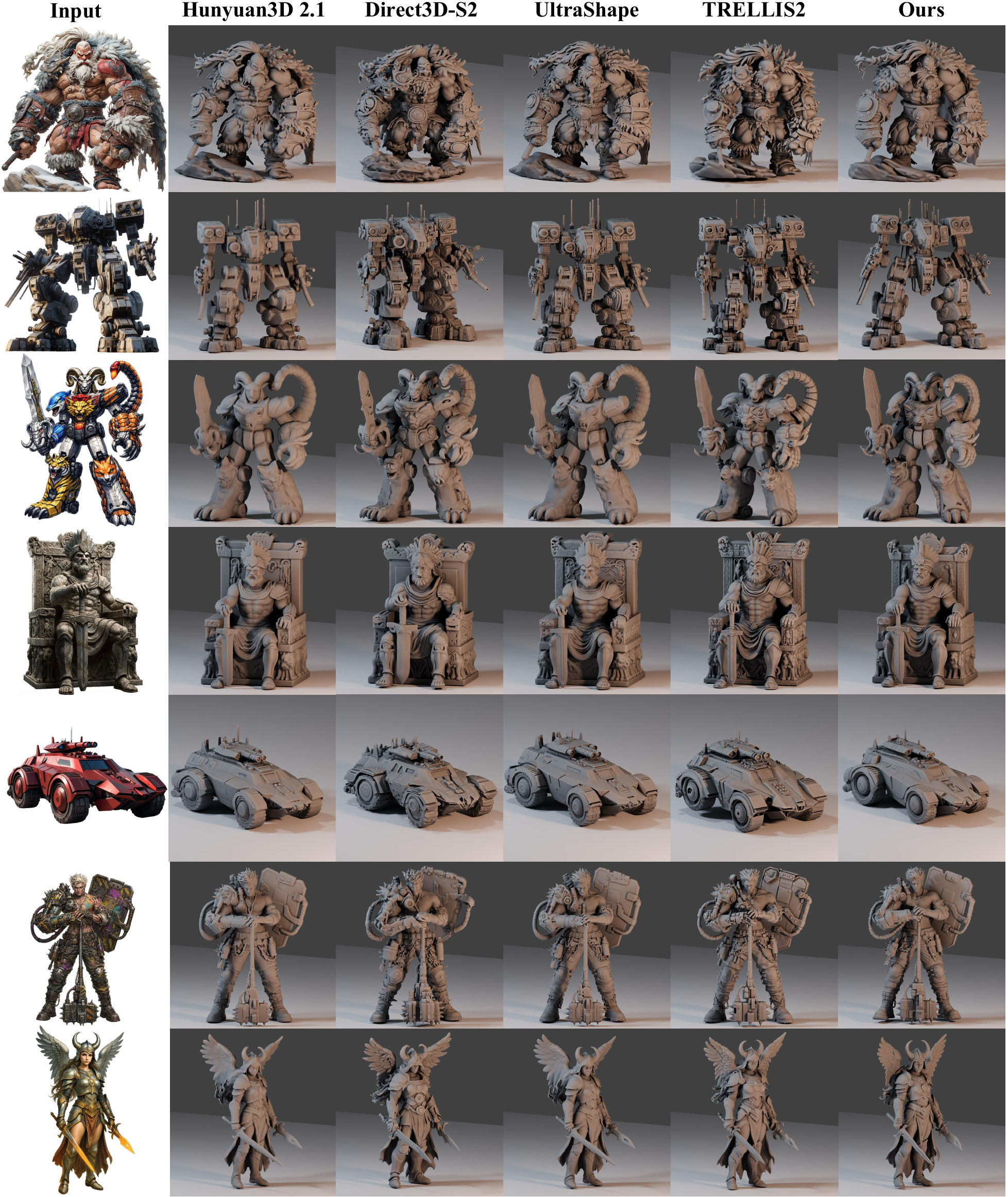}
  \caption{Additional qualitative comparisons with recent image-to-3D generation methods.}
  \Description{Additional qualitative comparisons for image-to-3D generation, showing the input images and generated meshes from different methods.}
  \label{fig:more_compare}
\end{figure*}

Figure~\ref{fig:more_compare} provides additional qualitative comparisons for image-to-3D generation. These examples complement the main results in Figure~\ref{fig:compare_gen} and include more diverse input images. Across these cases, Direct3D-V3 produces coherent complete shapes while preserving sharper local geometry and more regular surface details than competing methods.

\end{document}